\documentclass[10pt,journal,final]{IEEEtran}
\usepackage{times}
\usepackage{latexsym}
\usepackage{multirow}
\usepackage{graphicx}
\usepackage{makecell}
\usepackage{amsmath}
\usepackage{xcolor}
\usepackage{booktabs} 
\usepackage{array}
\usepackage{blindtext}
\usepackage{color}
\usepackage{colortbl}
\usepackage{multirow}
\usepackage{hyperref}
\usepackage{wrapfig}
\usepackage{caption}
\newcommand{\zqh}{\color{black}}
\newcommand{\minor}{\color{black}}
\newcommand{\our}{\textsc{PanDa}\xspace}

\usepackage{times}
\usepackage{latexsym}
\usepackage{multirow}
\usepackage{url}
\usepackage{latexsym}
\usepackage{balance}
\usepackage{bm}
\usepackage{amssymb}
\usepackage{amsmath}
\usepackage{url}
\usepackage{makecell}
\usepackage{multirow}
\usepackage{subfig}
\usepackage{graphicx} 
\usepackage{color}
\usepackage{colortbl}
\usepackage{textcomp}
\usepackage{CJK}
\usepackage{enumitem}
\usepackage{booktabs}
\usepackage{microtype}
\usepackage{arydshln}

\usepackage{pifont}
\usepackage[T1]{fontenc}
\usepackage[utf8]{inputenc}
\usepackage{inconsolata}

\begin{document}
%
\title{\our: Prompt Transfer Meets Knowledge Distillation for Efficient Model Adaptation}
%
%
%

\author{
        Qihuang~Zhong,~\IEEEmembership{Member,~IEEE,}
        Liang~Ding,~\IEEEmembership{Member,~IEEE,}
        Juhua~Liu,~\IEEEmembership{Member,~IEEE,}
        Bo~Du,~\IEEEmembership{Senior~Member,~IEEE,}
        and~Dacheng~Tao,~\IEEEmembership{Fellow,~IEEE}

\thanks{This work was supported in part by the National Key Research and Development Program of China under Grant 2023YFC2705700, and in part by the National Natural Science Foundation of China under Grant U23B2048, 62076186 and 62225113. The numerical calculations in this paper have been done on the supercomputing system in the Supercomputing Center of Wuhan University. \textit{Corresponding Author: Juhua Liu, Bo Du (e-mail: \{liujuhua, dubo\}@whu.edu.cn).}}

\thanks{Q. Zhong, J. Liu and B. Du are with the School of Computer Science, National Engineering Research Center for Multimedia Software, Institute of Artificial Intelligence, and Hubei Key Laboratory of Multimedia and Network Communication Engineering, Wuhan University, Wuhan, China (e-mail: \{zhongqihuang, liujuhua, dubo\}@whu.edu.cn).}

\thanks{L. Ding is with the School of Computer Science, Faculty of Engineering, The University of Sydney, Sydney, Australia (e-mail: liangding.liam@gmail.com).}

\thanks{D. Tao is with the College of Computing \& Data Science at Nanyang Technological University, \#32 Block N4 \#02a-014, 50 Nanyang Avenue, Singapore 639798 (e-mail: dacheng.tao@ntu.edu.sg).}
}

\maketitle


\begin{abstract}
    Prompt Transfer (PoT) is a recently-proposed approach to improve prompt-tuning, by initializing the target prompt with the existing prompt trained on similar source tasks. However, such a vanilla PoT approach usually achieves sub-optimal performance, as (i) the PoT is sensitive to the similarity of source-target pair and (ii) directly fine-tuning the prompt initialized with source prompt on target task might lead to forgetting of the useful general knowledge learned from source task. To tackle these issues, we propose a new metric to accurately predict the prompt transferability (regarding (i)), and a novel PoT approach (namely \our) that leverages the knowledge distillation technique to alleviate the knowledge forgetting effectively (regarding (ii)). Extensive and systematic experiments on 189 combinations of 21 source and 9 target datasets across 5 scales of PLMs demonstrate that: 1) \textit{our proposed metric works well to predict the prompt transferability}; 2) \textit{our \our consistently outperforms the vanilla PoT approach by 2.3\% average score (up to 24.1\%) among all tasks and model sizes}; 3) \textit{with our \our approach, prompt-tuning can achieve competitive and even better performance than model-tuning in various PLM scales scenarios}. We have publicly released our code in \url{https://github.com/WHU-ZQH/PANDA}.
\end{abstract}


\section{Introduction}
\label{introduction}
\IEEEPARstart{F}{}ine-tuning the pretrained language models (PLMs)~\cite{devlin2019bert,liu2019roberta,he2020deberta,raffel2020exploring,brown2020language} has become the \textit{de facto} standard for natural language processing (NLP). The dominant fine-tuning manner, \textit{i.e.,} model-tuning, is to tune the entire pretrained model parameters for each downstream task, which is computationally expensive and memory intensive~\cite{yuan2020parameter}. Hence, various parameter-efficient fine-tuning approaches are further explored~\cite{hu2021lora,lester2021power,li2021prefix}, among which \textit{prompt-tuning} has attracted great attention. Specifically, prompt-tuning refers to only tuning the soft prompt which is a set of trainable parameters added to the PLM, while keeping the PLM fixed. 

\begin{figure}[t]
	\centering
	\includegraphics[width=0.5\textwidth]{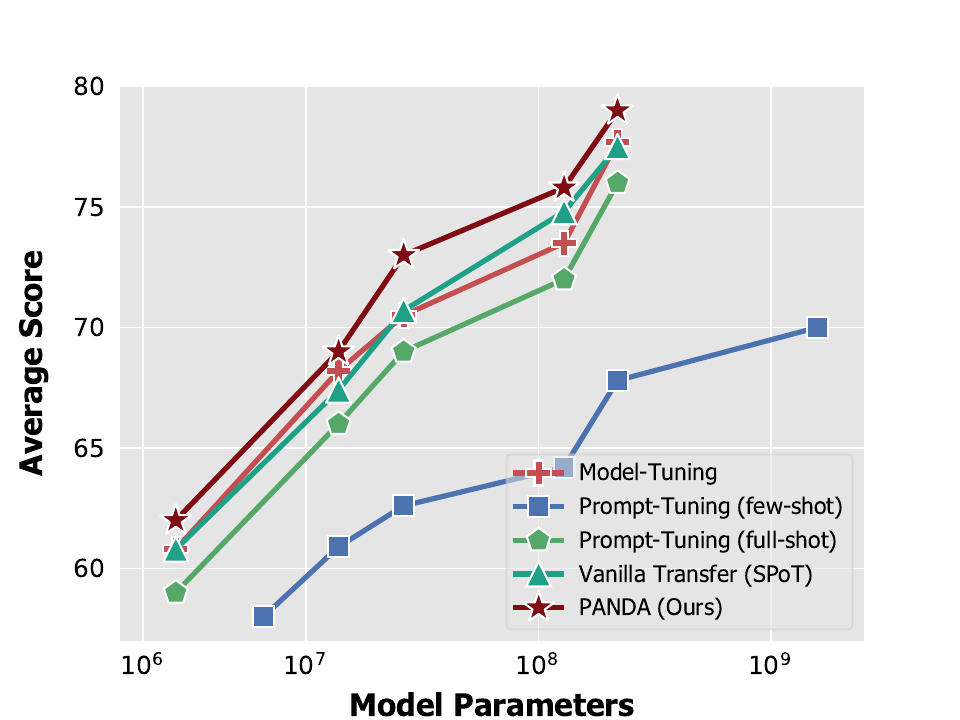} 
	\caption{Average performances on parts of SuperGLUE and GLUE benchmarks. Note that best performances of SPoT and our \our are reported. Our \our approach outperforms the vanilla prompt transfer approach (SPoT~\cite{vu2021spot}) across all model sizes. Additionally, within our \our approach, prompt-tuning method can obtain competitive or better performances than model-tuning methods in the full-shot scenario.}
	\label{fig:parameters}
\end{figure}

Prompt-tuning can achieve competitive performance against model-tuning when the PLM exceeds billions of parameters~\cite{lester2021power}, but there are still some gaps between prompt-tuning and model-tuning at smaller PLM scales~\cite{liu2021gpt, li2021prefix}, which can also be observed from our empirical results in Figure~\ref{fig:parameters}. Hence, it is crucial to explore how to boost the performance of prompt-tuning across all scales of PLMs. Intuitively, a consensus is that transferring knowledge from an intermediate task to the target task can improve the target performance~\cite{pruksachatkun2020intermediate}. Inspired by this, recent works~\cite{vu2021spot,su2021transferability} involve leveraging transfer learning in the context of prompt-tuning, \textit{i.e.}, \textit{prompt transfer} (PoT) that initializes the target prompt with the trained prompt of similar source tasks and then tunes the trained prompt on the target task. 
Although such a vanilla PoT approach can offer some improvements over prompt-tuning, it still has some limitations.
Specifically, the first is \ding{182} \textbf{\textit{
the performance of PoT is sensitive to the similarity between the source and target tasks}}. PoT highly relies on the prompt transferability metric to retrieve the useful source tasks for a given target task, but we empirically find that the prior metrics~\cite{vu2021spot,su2021transferability} fall short in distinguishing the task relationships and would hardly make a principled choice about which source tasks to use. 
Furthermore, given a similar source task, the vanilla PoT might still perform sub-optimal, as the second problem of \ding{183} \textbf{\textit{directly tuning the prompt initialized with source prompt on target task may lead to forgetting of the useful general knowledge learned from source task}}, which is similar to the conclusion in the existing work~\cite{chen2020recall} that sequential transfer learning of PLMs is prone to forget previously learned knowledge.

To this end, we first propose a new metric to better predict the prompt transferability, and then improve the \textbf{P}rompt tr\textbf{\textsc{a}}nsfer via k\textbf{\textsc{n}}owledge \textbf{D}istill\textbf{\textsc{a}}tion (\textbf{\our} for short). Specifically, for \ding{182}, different from the prior metric~\cite{vu2021spot} that simply uses the similarity of prompt parameters as prompt transferability, our proposed metric first maps the source/target tasks into a shared semantic space to obtain the task embeddings based on the source/target prompts, and then measures the prompt transferability by the similarity of corresponding task embeddings. On the other hand, as for \ding{183}, \our introduces the knowledge distillation technique to transfer the knowledge from source prompts to the target prompt in a subtle manner, thus alleviating the problem of prior knowledge forgetting effectively. 
Furthermore, to adaptively control the knowledge transfer in our \our approach, we use the prompt similarity predicted by our metric as the balancing factor between two supervision signals for each source-target pair.

We conduct a large-scale study using the 189 combinations of 21 source and 9 target datasets across 5 scales of PLMs. Qualitative and quantitative analyses show that our proposed metric not only performs better to distinguish different task relationships, but also aligns with the transfer performance accurately. Moreover, results of PoT demonstrate that our \our achieves significant improvements over vanilla PoT across all tasks and model sizes (up to 24.1\% average scores in some scenarios), and makes prompt-tuning obtain competitive and even better performance than model-tuning in the full-data scenario.

To summarize, \textbf{our contributions} are as follows: 
\begin{itemize}
    \item We recast vanilla PoT with \our, a novel prompt transfer approach, which leverages knowledge distillation to alleviate the prior-learn knowledge forgetting problem.
    \item We propose a new prompt transferability metric to wisely choose the more helpful source tasks, for a given target task.
    \item Extensive and systematic experiments on 189 pairs of source-target tasks across 5 scales of PLMs prove the effectiveness of our methods.
\end{itemize}

The rest of this paper is organized as follows. In Section~\ref{sec_related}, we briefly review the related works. In Section~\ref{sec_method}, we introduce our proposed framework in detail. Section~\ref{sec_experiment} reports and analyzes our experimental results, followed by the discussion in Section~\ref{sec_discussion}. Lastly, we conclude our study in Section~\ref{conclusion}.
\section{Related Works}
\label{sec_related}

\subsection{Pretrained Language Model}
In recent years, we have witnessed numerous Transformer~\cite{transformer}-based pretrained language models (PLMs)~\cite{devlin2019bert,liu2019roberta,he2020deberta,joshi2020spanbert,brown2020language,raffel2020exploring,lewis2020bart,zhong2022toward,zhong2023bag,zhong2023self} that achieved tremendous success in various natural language understanding (NLU) and generation (NLG) tasks. Based on the model architectures, these PLMs can be classified into three groups: 1) decoder-only PLMs (\textit{e.g.}, GPT-3~\cite{brown2020language}), 2) encoder-only PLMs (\textit{e.g.}, BERT~\cite{devlin2019bert}), and 3) encoder-decoder PLMs (\textit{e.g.}, T5~\cite{raffel2020exploring}). 

Due to different pretraining functions, these PLMs exhibit different abilities when performing NLP tasks. Specifically, decoder-only PLMs generally adopt the auto-regressive language modeling objectives, which aim to predict the future words towards a sequence of words~\cite{brown2020language}. The representative decoder-only PLMs are GPT-3~\cite{brown2020language} and its variants. Such auto-regressive models are well-suitable for language generation tasks, but they are unidirectional and usually {\zqh fall} short in the representation learning for understanding the sentence~\cite{liu2021pre}. Thus, researchers turn to focus on discriminative (encoder-only) PLMs that introduce a bidirectional masked language modeling (MLM) objective~\cite{devlin2019bert} to predict the masked text token based on the context. The most representative encoder-only PLMs are BERT~\cite{devlin2019bert} and its variants, \textit{e.g.}, RoBERTa~\cite{liu2019roberta} and DeBERTa~\cite{he2020deberta}. 
Additionally, to combine the advantages of decoder-only PLMs and encoder-only PLMs, encoder-decoder PLMs are sequentially proposed (\textit{e.g.}, T5~\cite{raffel2020exploring} and BART~\cite{lewis2020bart}), which firstly employ a separate encoder to model the source text and then use a left-to-right LM to decode the conditional target text. The encoder-decoder paradigm makes these PLMs not only generally suitable for text generation, but also well for text understanding tasks~\cite{zhong2023e2s2}.

In this paper, we mainly focus on the adaptation of discriminative (encoder-only) language models and aim to improve its efficiency via a simple yet effective prompt-tuning method. Hence, we review the related work of prompt-tuning on discriminative language models in the following part.

\subsection{Prompt-tuning for PLMs}
As aforementioned above, PLMs have achieved tremendous success in the community of NLP~\cite{guan2020deep,zhong2022knowledge,li2020few,chen2021user,li2020neural,li2020survey,zhong2022improving,zhong2023revisiting}.
However, as the model size continues increasing, it becomes more costly and impractical to apply these large-scale PLMs to downstream tasks, as the current dominant model-tuning needs to tune all pretrained parameters for each task.  

To address this issue, researchers turn to focus on other parameter-efficient methods~\cite{hu2021lora,xu2021raise,houlsby2019parameter,huang2023event} for efficiently fine-tuning the PLMs. These works tend to only update small parts of the language model while keeping the other pre-trained parameters fixed~\cite{xu2021raise}, or design and train some task-specific modules (\textit{e.g.}, adapters~\cite{houlsby2019parameter} and low-rank structures~\cite{hu2021lora}) for each downstream task. Among these methods, \textit{prompt-tuning}, which only tunes the soft prompt that is a set of trainable embeddings/parameters added into the frozen PLMs, has attracted great attention recently~\cite{li2021prefix,liu2021p,liu2021gpt,gu2021ppt}. Notably, earlier works focus on exploring the discrete prompt, such as manually-designed prompt~\cite{schick2020few,schick2021exploiting} and automatically-searched prompt~\cite{shin2020autoprompt,han2021ptr}, which are almost hard tokens and sensitive to the prompt itself, \textit{i.e.}, minor changes in the prompt might lead to significantly different performance. Thus, more recent efforts attempt to investigate soft prompt~\cite{li2021prefix,liu2021p,liu2021gpt,gu2021ppt} that can be updated with task-specific supervision. Our work aims to analyze the effect of soft prompt, thus we refer to the prompt-tuning as that with soft prompts in this paper, instead of discrete hard prompts.

\begin{figure}[t]
	\centering
	\includegraphics[width=0.5\textwidth]{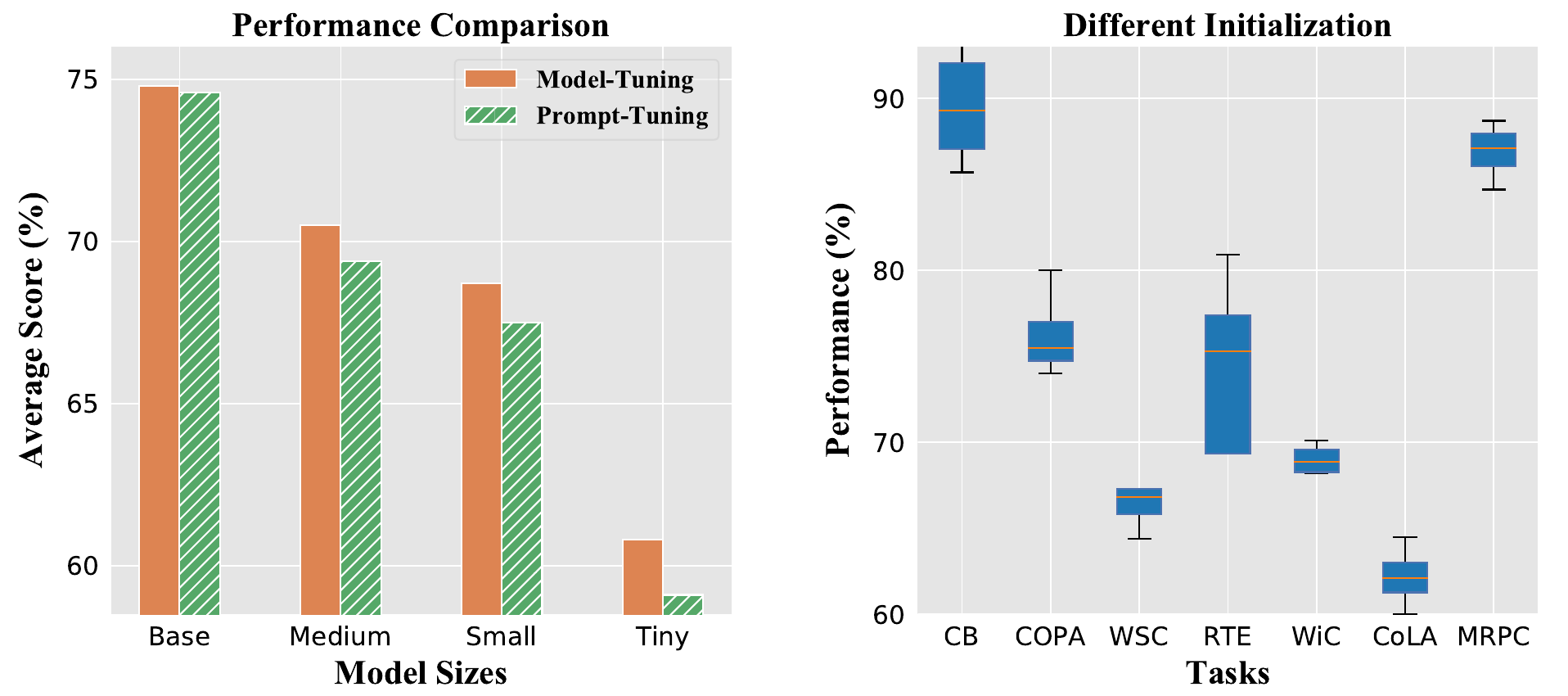} 
	\caption{~~\textbf{Left}: Comparisons between model-tuning and prompt-tuning across various model sizes. {\zqh Here, we report the average performance among all 9 target tasks (as stated in Table~\ref{tab:dataset_details}).} \textbf{Right}: Sensitivity analysis of prompt-tuning on the initialization of prompt, where the normal/sparse/constant (all zeros) initialization is used. {\zqh BERT-large is used in this setting.}}
\label{fig:challenge}
\end{figure}

Lester~et~al.~\cite{lester2021power} show that prompt-tuning can achieve comparable performance to model-tuning in the full-data scenario when the PLM has billions of parameters.
However, the performance of prompt-tuning on smaller PLMs is still sub-optimal~\cite{li2021prefix,gu2021ppt} and significantly sensitive to the prompt initialization~\cite{lester2021power}, which can also be empirically proved by our preliminary experiments in Figure~\ref{fig:challenge}. Hence, PoT~\cite{vu2021spot,su2021transferability, asai2022attempt, peng2022model} methods (\textit{e.g.}, SPoT~\cite{vu2021spot}, the first of PoT works) are further proposed, which first learn soft prompts on source tasks and then use them to initialize the prompt for a target task, to improve the prompt-tuning. {\zqh We refer to these ``\textit{training and then directly tuning}'' PoT methods as \textbf{Vanilla PoT} methods.
Upon the vanilla PoT, SPoT~\cite{vu2021spot} and Su~et~al.~\cite{su2021transferability} additionally explore the transferability metrics to predict the optimal source task. ATTEMPT~\cite{asai2022attempt} and SESoM~\cite{peng2022model} further expand the vanilla PoT methods to the multi-task settings, \textit{i.e.}, transferring the knowledge from multiple source tasks to the target prompt.}

\subsection{Knowledge Distillation} 
Knowledge distillation (KD), which aims to extract the ``dark knowledge'' from the teacher network to guide the training of the student network, has emerged as an important technique for transfer learning and model compression~\cite{hinton2015distilling,xu2020knowledge,furlanello2018born,zhong2024revisiting}. Specifically, the student network is trained with the supervision signals from both ground-truth labels and the teacher network. The common way to leverage the supervision information of teacher model is to match the outputs of teacher and student models by minimizing the distance of the output distribution~\cite{hinton2015distilling}. KD has the powerful ability to transfer knowledge, which has been proved in many fields~\cite{ding2021understanding,hahn2019self}. Thus, an intuitive idea is to employ the KD to tackle the problem of knowledge forgetting in PoT. To this end, we introduce the KD into PoT for better knowledge transfer and to boost the performance of prompt-tuning. 

 \noindent \textbf{{\zqh Clarification of novelty. }} Some readers may concern that our work is similar to the prior PoT methods, \textit{i.e.}, SPoT~\cite{vu2021spot}. Here, we depart from the SPoT and ours as follows: 1) \textit{different motivations}: instead of verifying the effect of vanilla PoT on prompt-tuning, we aim to alleviate the knowledge forgetting problem of PoT by leveraging the knowledge distillation for a subtle knowledge transfer; 2) \textit{different methods}: in addition to \our framework, we also propose a novel metric to i) accurately estimate the prompt transferability and ii) adaptively control the knowledge transfer in \our, which is significantly different from that in SPoT; 3) \textit{more analyses}: we provide more experimental results and analyses towards the effectiveness of our methods in more complex scenarios, {\zqh \textit{e.g.}, multi-task PoT}.

\section{Method}
\label{sec_method}
We present the preliminaries of prompt-tuning and PoT, the new metric to predict prompt transferability between source and target tasks, and our proposed \our approach.
\subsection{Preliminaries}
In this subsection, we introduce the background of model-tuning and prompt-tuning. Suppose that we focus on fine-tuning the discriminative PLMs trained with masked language modeling objective~\cite{devlin2019bert}. Let $\mathcal{M}$ denote a PLM and $\boldsymbol{x}$ be a sequence of input tokens. Firstly, $\boldsymbol{x}$ will be converted to a fixed token sequence $\tilde{\boldsymbol{x}}$ = \texttt{[CLS]}$\boldsymbol{x}$\texttt{[SEP]}, where \texttt{[CLS]} is a special token used to capture the global information and \texttt{[SEP]} is used to separate the input sequences. Then, $\tilde{\boldsymbol{x}}$ is sequentially mapped to a sequence of hidden vector $\mathbf{h}=\mathcal{M}(\tilde{\boldsymbol{x}}) \in \mathbb{R}^d$ via the $\mathcal{M}$. 
Lastly, the conventional model-tuning trains a classification head $f$ (\textit{e.g.}, an MLP layer) on top of model outputs $\mathbf{h}$. In practice, the outputs of \texttt{[CLS]} token (\textit{i.e.}, $h_{cls}$) are usually used to represent the input sequence and fed into $f$ to predict an output class.
Specifically,
\begin{align*}
    \texttt{[CLS]} \boldsymbol{x} \texttt{[SEP]} \xrightarrow{\mathcal{M}} \mathbf{h} \xrightarrow{f(h_{cls})} P(y|\boldsymbol{x};\mathcal{M},f) ,
    \\
    \mathcal{L}_{ce}(\mathcal{M}, f) = -\sum_{i\in \mathcal{N}} \log P( y_i | \boldsymbol{x}_i;\mathcal{M}, f ),
\end{align*}
where $y$ denotes the target label and $\mathcal{N}$ is the number of training samples. Model-tuning aims to minimize the classification loss $\mathcal{L}_{ce}(\mathcal{M}, f)$ on the set of all parameters of $\mathcal{M}$ and $f$.

Differently, prompt-tuning fixes $\mathcal{M}$ and introduces the soft prompt (some trainable parameters) for leveraging the fixed $\mathcal{M}$ to predict outputs. In practice, letting $e(\tilde{\boldsymbol{x}})$ denote the token embeddings of $\tilde{\boldsymbol{x}}$, prompt-tuning directly concatenates a trainable soft prompt $u$ to the token embeddings~\cite{li2021prefix} or add $u$ in each transformer layer~\cite{liu2021p}, which can be uniformly denoted as $(u, e(\tilde{\boldsymbol{x}}))$. Then, we view the $\mathcal{M}$ as a function of token embeddings and refer this model by $\overline{\mathcal{M}}$, \textit{i.e.}, $\overline{\mathcal{M}}(e(\tilde{\boldsymbol{x}}))$ = $\mathcal{M}(\tilde{\boldsymbol{x}})$. In this way, the model outputs of prompt-tuning are $\overline{\mathcal{M}}(u,e(\tilde{\boldsymbol{x}}))$ and we consider simultaneously training the prompt parameter $u$ and a classification head $f$ via the classification loss $\mathcal{L}_{ce}(u, f)$:
\begin{equation}
    \mathcal{L}_{ce}(u, f) = -\sum_{i\in \mathcal{N}} \log P ( y_i | e(\tilde{\boldsymbol{x}}_i);\overline{\mathcal{M}}, u, f ).
    \label{eq_ce} 
\end{equation}
Note that prompt-tuning minimizes the classification loss $\mathcal{L}_{cl}(u, f)$ and only updates parameters of $u$ and $f$ while keeping $\overline{\mathcal{M}}$ fixed.

\begin{figure*}[ht]
	\centering
	\includegraphics[width=0.95\textwidth]{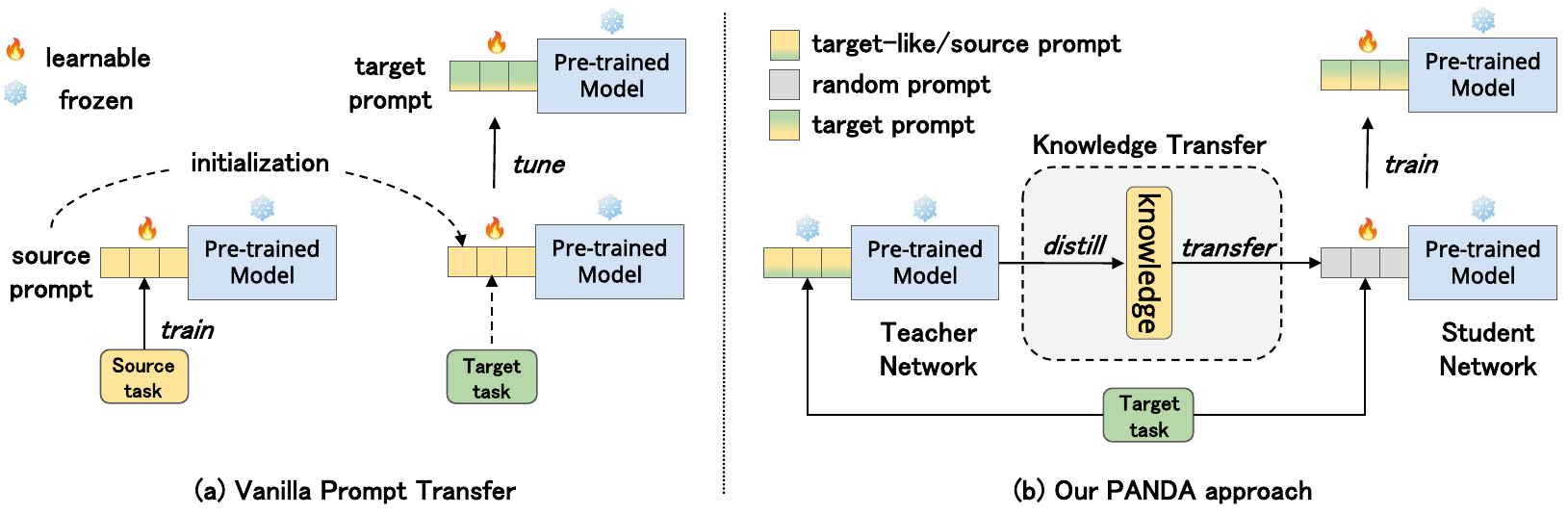} 
	\caption{\textbf{Left}: An illustration of vanilla PoT. \textbf{Right}: The architecture of our proposed PANDA. Notably, we can first train the teacher network on the target task with fewer iterations and obtain the new teacher network (target-like network), but we do not show the procedure for ease of illustration.
	}
	\label{fig:panda}
\end{figure*}

\subsection{Prompt transfer}
Prompt-tuning is a parameter-efficient approach and shows competitive performance against model-tuning in large-scale PLMs, but raises two challenges. The first is the performance drop with the decrease of PLM scales, as shown in Figure~\ref{fig:challenge} (Left), prompt-tuning performs much worse than model-tuning when the PLM is smaller. Another challenge is that prompt-tuning is sensitive to the initialization of prompt parameters~\cite{lester2021power}. Figure~\ref{fig:challenge} (Right) shows the results of the prompt initialized with different manners, {\zqh \textit{i.e.}, different parameter initialization methods (such as normal, sparse and constant initialization functions implemented in Pytorch),} on various tasks.

Regrading these challenges, Vu~\textit{et~al.}~\cite{vu2021spot} propose a new transfer learning approach in the context of prompt-tuning, \textit{i.e.}, prompt transfer (PoT). As illustrated in Figure~\ref{fig:panda} (Left), vanilla PoT first trains a soft prompt $u_s$ on one source task and then uses the trained prompt to initialize the prompt for a target task. Lastly, the prompt initialized with $u_s$ is further fine-tuned on the target task to obtain the task-specific target prompt $u_t$.

\subsection{Proposed Methods}
\label{sec_panda}
Here, we first propose a new metric to predict prompt transferability and then introduce the details of our \our approach.

\subsubsection{Prompt Transferability Metric}
Inspired by prior works~\cite{gururangan2020don, vu2021spot,jiang2022promptbert}, we attempt to construct a semantic space of tasks based on soft prompt and map source-target tasks to this vector space. Then, we can measure a cosine similarity between the corresponding task embeddings and use this similarity as prompt transferability. In practice, {\zqh to avoid domain and label biases, we randomly select parts of samples (denoted as $\mathcal{D}$) for each class, \textit{e.g.}, 50 samples with positive labels and 50 samples with negative labels for the sentiment analysis task} (the detailed analysis of the number can be found in Section~\ref{ablation_sample}), from each task as the representative data for fast computation. Then, the data are respectively fed into the original PLM $\mathcal{M}$ and the PLM with trained prompt $[\overline{\mathcal{M}};u]$ to obtain the hidden vectors of \texttt{[CLS]}, \textit{i.e.}, $\mathcal{D} \xrightarrow[\texttt{[CLS]}]{\mathcal{M}} h_{cls}^m$ and $e(\mathcal{D}) \xrightarrow[\texttt{[CLS]}]{\overline{\mathcal{M}},\,u} h_{cls}^p$. To exactly represent the effect of soft prompt, we follow Jiang~et~al.~\cite{jiang2022promptbert} and use the $\hat{h}=h_{cls}^p-h_{cls}^m$ as the prompt-based task embeddings for reducing the influence of PLM itself\footnote{As the parameter number of the soft prompt is much smaller than that of PLM itself, the representation difference of different tasks might be more influenced by PLM, which would affect the prediction accuracy.}. In this way, we can obtain the corresponding embeddings $\hat{h}_s$ and $\hat{h}_t$ for source and target tasks, which are then used to measure the similarity\footnote{{\zqh Following many previous works~\cite{vu2021spot,su2021transferability,wang2022contrastive}, we use the representative cosine similarity to measure the similarity. Exploring more sophisticated similarity metrics (\textit{e.g.}, Manhattan Distance and Jaccard Similarity) may improve the accuracy, but it is not our main focus in this paper.}} $sim(\hat{h}_s,\hat{h}_t)$ as:
\begin{equation}
\begin{aligned}
sim(\hat{h}_s,\hat{h}_t) = (\frac{\hat{h}_s \cdot \hat{h}_t}{\|\hat{h}_s\| \|\hat{h}_t\|}).
\label{eq_metric}
\end{aligned}
\end{equation}

\subsubsection{\our Approach}
As illustrated in Figure~\ref{fig:panda} (Right), our \our approach introduces the knowledge distillation technique to facilitate the knowledge transfer from the source task to the target task. In particular, \our approach first uses the PLM with source prompt as the teacher network $[\overline{\mathcal{M}};u_s]$, and the PLM with randomly initialized prompt (denoted as $u_r$) as the student network $[\overline{\mathcal{M}};u_r]$. 
Notably, we can then train $[\overline{\mathcal{M}};u_s]$ on the target task with fewer iterations {\zqh (\textit{i.e.}, 1 epoch in the small-scale tasks (\textit{e.g.}, CB) or 1000 fixed steps in the larger tasks (\textit{e.g.}, MNLI), where the detailed task information can be found in Table~\ref{tab:dataset_details})} to obtain the new teacher (denoted as target-like network $[\overline{\mathcal{M}};u^*_s]$), which learns some target information but not forget the source knowledge significantly. We state that such a target-like network performs as an intermediate network to bridge the gap between source and target tasks\footnote{We empirically find that using the original teacher network (\textit{i.e.}, $[\overline{\mathcal{M}};u_s]$) could achieve comparable performance than the target-like network (see in Section~\ref{teacher_ablation}), \textit{i.e.}, we can simply employ the source prompt itself as the teacher prompt, for the novel target tasks required fast computation.}.
Sequentially, the student network is trained on the target task with the supervision information of both ground-truth labels (same as Equation~\ref{eq_ce}), and soft targets predicted by the teacher network that can be formulated as\footnote{Note that we simplify $P(y_i | e(\tilde{\boldsymbol{x}}_i);\overline{\mathcal{M}}, u, f )$ of Equation~\ref{eq_ce} to $P ( y_i | u)$ here. In the preliminary experiments, we found that such an MSE loss achieved the best performance in most settings, and thus we used it lastly. The analysis of different KD losses can be seen in Section~\ref{ablation_loss}.}:
\begin{equation}
    \mathcal{L}_{kd}(u_r, f) =
    \sum_{i\in \mathcal{D}} 
    ( P_{te} ( y_i | u^*_s) - P_{st} ( y_i | u_r))^2,  
    \label{eq_kd}
\end{equation}
where $P_{te}$ and $P_{st}$ denote the distributions predicted by the teacher and student networks respectively. Forcing the student to mimic the prediction of teacher can make the student learn the ``dark knowledge'' from the teacher, thus guiding the knowledge transfer from source task to target prompt. In summary, the complete loss function $\mathcal{L}_{all}$ of student network can be formulated as:
\begin{equation}
    \mathcal{L}_{all}(u_r, f) = 
    \mathcal{L}_{ce}(u_r, f) + \lambda \cdot sim\cdot \mathcal{L}_{kd}(u_r, f), 
    \label{eq_beta} 
\end{equation}
where $sim(\hat{h}_s,\hat{h}_t)$ in Equation~\ref{eq_metric} ($sim$ for short) is used to adaptively control the knowledge transfer, and $\lambda$ is a factor to adjust the maximum of transfer ratio, which is empirically set as 0.05 (we analyze the influence of different $\lambda$ in Section~\ref{ablation_factor}).

\subsubsection{Expanding to multi-task settings}
In this part, we attempt to expand our \our approach to multi-task settings, i.e., transferring the knowledge of multiple source tasks into a given target task. 
Specifically, suppose that we have $n$ source (or target-like\footnote{As aforementioned, we can further obtain the target-like prompts based on source prompts. For ease of understanding, we uniformly use source prompts for description here.}) prompts $U_s=\{u_{s_1},u_{s_2},...,u_{s_n}\}$ and corresponding transferability $\{sim_1,sim_2,...,sim_n\}$, the key problem is how to fuse the knowledge of them.
Inspired by the prior multi-task studies~\cite{asai2022attempt,peng2022model,zhong2022knowledge}, we employ a early-fusion strategy and a late-fusion strategy to fuse the multi-source knowledge, respectively. 
Regrading the early-fusion strategy, we first directly fuse multiple soft prompts as an ensemble teacher prompt and then use it to guide the training of student prompt, as done in Equation~\ref{eq_kd} and~\ref{eq_beta}.
Specifically, the process of early-fusion can be formulated as following:
\begin{equation}
    \begin{split}
    &u_{s_{all}} = \frac{1}{n} \sum_{i}^n sim_i * u_{s_i} \\
    &\mathcal{L}_{kd_{early}}(u_r, f) =
    \sum_{i\in \mathcal{D}} 
    ( P_{te} ( y_i | u_{s_{all}}) - P_{st} ( y_i | u_r))^2,\\
    &\mathcal{L}_{all_{early}}(u_r, f) = 
    \mathcal{L}_{ce}(u_r, f) + \lambda \cdot \mathcal{L}_{kd_{early}}(u_r, f),
    \end{split}
    \label{early_fusion} 
\end{equation}
{\zqh where $u_{s_{all}}$ is the weighted average of source prompts $\{u_{s_1},u_{s_2},...,u_{s_n}\}$, and we denote the early-fusion KD loss function as $\mathcal{L}_{kd_{early}}$. Notably, the prompt similarities are used to fuse the sources prompts, we do not use them to explicitly control the weight of $\mathcal{L}_{kd_{early}}$.}

For the late-fusion strategy, we first obtain the multiple teacher representations and use the ensemble representation as the supervision information for KD training. Specifically, this process can be formulated as:
\begin{equation}
    \begin{split}
    &P_{te_{all}} ( y_i | U_s) = \frac{1}{n} \sum_{i}^n sim_i * P_{te} ( y_i | u_{s_i}),\\
    &\mathcal{L}_{kd_{late}}(u_r, f) =
    \sum_{i\in \mathcal{D}} 
    ( P_{te_{all}} ( y_i | U_s) - P_{st} ( y_i | u_r))^2,\\
    &\mathcal{L}_{all_{late}}(u_r, f) = 
    \mathcal{L}_{ce}(u_r, f) + \lambda \cdot \mathcal{L}_{kd_{late}}(u_r, f),
    \end{split}
    \label{early_fusion} 
\end{equation}
{\zqh where $P_{te_{all}}$ is the weighted average of teacher representations, and we denote the late-fusion KD loss function as $\mathcal{L}_{kd_{late}}$.}

Through fusing the knowledge of multiple source tasks, we can easily obtain the general knowledge and transfer it into the target prompt by \our. It should also be noted that our main contribution is to prove that PoT can be further improved by a KD technique, but not to design a more effective KD method. Thus, we simply use a vanilla KD in this work, and we believe that a more sophisticated KD can achieve further improvements in both single-task and multiple-task PoT scenarios.

\begin{table*}[t]
\centering
\setlength{\tabcolsep}{14pt}
\scalebox{1}{
\begin{tabular}{cccccccc}
\toprule
\multicolumn{1}{l}{\multirow{2}{*}{}} &
 \multirow{2}{*}{\textbf{Name}} &
 \multirow{2}{*}{\textbf{Task Type}} &
 \multirow{2}{*}{\textbf{Tranfer Type}} &
 \multirow{2}{*}{\textbf{|Train|}} &
 \multicolumn{3}{c}{\textbf{Training Details}} \\ \cline{6-8}
\multicolumn{1}{l}{}                  &
                       & 
                            &
                               &
                          &
 \textbf{lr}    &
 \textbf{bsz}   &
 \textbf{epoch}   \\ \hline \hline
\multirow{8}{*}{SuperGLUE}            &
 Boolq                 &
 Question Answering                         &
 source                        &
 9K                       &
 5e-3             &
 32           &
 100     \\
                                      &
 CB                    &
 Natural Language Inference                        &
 source/target                 &
 250                      &
 1e-2             &
 16           &
 100     \\
                                      &
 RTE                   &
 Natural Language Inference                        &
 source/target                 &
 2K                       &
 1e-2             &
 32           &
 60      \\
                                      &
 WIC                   &
 Word Sense Disambiguation  &
 source/target                 &
 5K                       &
 1e-2             &
 16           &
 80      \\
                                      &
 WSC                   &
 Natural Language Inference     &
 source/target                 &
 554                      &
 5e-3             &
 16           &
 80      \\
                                      &
 COPA                  &
 Question Answering                         &
 source/target                 &
 400                      &
 1e-2             &
 16           &
 80      \\
                                      &
 Multirc               &
 Question Answering                         &
 source                        &
 27K                      &
 5e-3             &
 32           &
 40      \\
                                      &
 Record                &
 Question Answering                         &
 source                        &
 25K                      &
 5e-3             &
 16           &
 5       \\ \midrule
\multirow{7}{*}{GLUE}                 &
 MNLI                  &
 Natural Language Inference                        &
 source                        &
 393K                     &
 5e-3             &
 64           &
 20      \\
                                      &
 CoLA                  &
 Grammatical Acceptability  &
 source/target                 &
 9K                       &
 5e-3             &
 16           &
 40      \\
                                      &
 SST2                  &
 Sentiment Analysis                         &
 source                        &
 67K                      &
 5e-3             &
 32           &
 60      \\
                                      &
 QNLI                  &
 Natural Language Inference                        &
 source                        &
 105K                     &
 5e-3             &
 32           &
 20      \\
                                      &
 MRPC                  &
 Paraphrase Detection       &
 source/target                 &
 4K                       &
 1e-2             &
 32           &
 60      \\
                                      &
 STSB                  &
 Semantic Similarity        &
 source/target                 &
 6K                       &
 1e-2             &
 32           &
 60      \\
                                      &
 QQP                   &
 Paraphrase Detection       &
 source                        &
 364K                     &
 5e-3             &
 64           &
 20      \\ \midrule
\multirow{6}{*}{Others}               &
 SQuAD                 &
 Question Answering                         &
 source                        &
 88K                      &
 5e-3             &
 8            &
 10      \\
                                      &
 CoNLL03               &
 Named Entity Recognition                        &
 source                        &
 14K                      &
 3e-2             &
 16           &
 30      \\
                                      &
 CoNLL04               &
 Named Entity Recognition                        &
 source/target                 &
 928                      &
 2e-2             &
 32           &
 40      \\
                                      &
 CoNLL05               &
 Semantic Role Labeling                        &
 source                        &
 90K                      &
 6e-3             &
 16           &
 15      \\
                                      &
 CoNLL12               &
 Semantic Role Labeling                        &
 source                        &
 82K                      &
 5e-3             &
 16           &
 45      \\
                                      &
 Ontonotes             &
 Named Entity Recognition                        &
 source                        &
 60K                      &
 1e-2             &
 16           &
 30     \\ \bottomrule
\end{tabular}
}
\caption{Details of all datasets that we study in this paper. Note that ``\textbf{|Train|}'' is  the training corpus size and ``\textbf{Training Details}'' are hyper-parameters of prompt tuning for each dataset. {\zqh Here, for clarification, we refer to the tasks containing less than 20K training samples as small tasks, while the others as large tasks.}}
\label{tab:dataset_details}
\end{table*}

\renewcommand\arraystretch{0.85}
\begin{table*}[]
\centering
\setlength{\tabcolsep}{10.5pt}
\scalebox{1}{
\begin{tabular}{lcccccccccl}
\toprule
\textbf{Tasks}    &
\textbf{CB}       & 
\textbf{COPA}     &
\textbf{WSC}      & 
\textbf{RTE}      & 
\textbf{WIC}      & 
\textbf{CoLA}     &
\textbf{MRPC}     &
\textbf{STSB}     & 
\textbf{CoNLL04}  & 
\textbf{Avg.}  
\\  \midrule \midrule
\zqh{model-tuning} & 
94.6 & 
69.0 & 
68.3 & 
75.8 & 
74.9 & 
60.6 & 
88.0 & 
90.0 & 
85.6 & 
78.5 
\\
\zqh{Lester et al.~\cite{lester2021power}}   &
\zqh{78.6}          &
\zqh{68.0}          &
\zqh{63.5}          &
\zqh{60.6}          &
\zqh{62.4}          &
\zqh{43.1}          &
\zqh{76.3}          &
\zqh{86.1}          &
\zqh{81.9}             &
\zqh{68.9}                                  \\
\zqh{P-Tuning-v2~\cite{liu2021p}}  & 
87.5 & 
76.0 & 
64.4 & 
76.2 & 
66.9 & 
63.8 & 
86.8 & 
90.5 & 
85.5 & 
77.5 
\\ \midrule

\multicolumn{11}{c}{\textbf{\textit{(a) Transfer with Vanilla Prompt Transfer approach~\cite{vu2021spot}.}} \zqh{P-Tuning-v2 is used for prompt-tuning in PoT.}}   \\
\midrule

BoolQ  & 
87.5   & 
68.0   & 
{\textbf{66.3}} & 
73.3   & 
65.0   & 
59.3   & 
86.5   & 
87.9   & 
{\textbf{85.6}} & 
\textbf{75.5}                        
\\
 
CB     & 
87.5   & 
74.0   & 
64.4   & 
73.3   &
66.0   & 
{\textbf{64.3}} & 
{\textbf{88.5}} & 
90.3   & 
85.0   & 
\textbf{77.0}                        
\\
 
RTE    & 
82.1   & 
{\textbf{79.0}} & 
65.4   & 
76.2   & 
{\textbf{67.9}} & 
62.1   & 
85.3   & 
90.2   & 
85.0   & 
\textbf{77.0}                        
\\
 
WIC    & 
85.7                                 & 
74.0                                 & 
{\textbf{65.4}} & 
{\textbf{76.5}} & 
66.9                                 & 
54.2                                 & 
84.6                                 & 
87.4                                 & 
84.4                                 & 
\textbf{75.5}                        
\\
 
WSC                                  & 
{\textbf{89.3}} & 
{\textbf{77.0}} & 
64.4                                 & 
73.6                                 & 
{\textbf{67.9}} & 
{\textbf{64.5}} & 
{\textbf{87.3}} & 
{\textbf{90.7}} & 
85.5                                 & 
{\textbf{77.8}} 
\\
 
COPA                                 & 
87.5                                 & 
76.0                                 & 
64.4                                 & 
76.2                                 & 
{\textbf{67.7}} & 
62.4                                 & 
86.3                                 & 
90.3                                 & 
84.2                                 & 
\textbf{77.2}                        
\\
 
MultiRC                              & 
87.5                                 & 
75.0                                 & 
{\textbf{65.4}} & 
74.4                                 & 
{\textbf{67.9}} & 
60.4                                 & 
{\textbf{88.0}} & 
90.3                                 & 
84.5                                 & 
\textbf{77.0}                        
\\
  
ReCoRD                               & 
78.6                                 & 
63.0                                 & 
{\textbf{65.4}} & 
53.8                                 & 
51.7                                 & 
0.0                                  & 
77.7                                 & 
85.0                                 & 
82.7                                 & 
\textbf{62.0}                        
\\ \midrule
 
MNLI                                 & 
{\textbf{96.4}} & 
71.0                                 & 
{\textbf{67.3}} & 
{\textbf{80.9}} &
66.5                                 & 
58.9                                 & 
{\textbf{88.2}} & 
{\textbf{91.0}} & 
83.0                                 & 
{\textbf{78.1}}  
\\
 
CoLA                                 & 
{\textbf{89.3}} & 
{\textbf{77.0}} & 
{\textbf{65.4}} & 
70.4                                 & 
{\textbf{67.1}} & 
63.8                                 & 
85.5                                 & 
90.0                                 & 
85.0                                 & 
\textbf{77.1}                        
\\
 
SST2                                 & 
{\textbf{92.9}} & 
74.0                                 & 
64.4                                 & 
71.8                                 & 
66.8                                 & 
60.1                                 & 
{\textbf{87.0}} & 
89.6                                 & 
84.3                                 & 
\textbf{76.8}                        
\\
 
QNLI                                 & 
{\textbf{89.3}} & 
76.0                                 & 
{\textbf{65.4}} & 
76.2                                 & 
{\textbf{70.4}} & 
63.7                                 & 
{\textbf{88.5}} & 
{\textbf{90.7}} & 
83.5                                 & 
{\textbf{78.2}}  
\\
 
MRPC                                 & 
{\textbf{89.3}} & 
{\textbf{78.0}} & 
{\textbf{67.3}} & 
74.4                                 & 
{\textbf{68.3}} & 
{\textbf{64.1}} & 
86.8                                 & 
90.3                                 & 
84.3                                 & 
{\textbf{78.1}}                       
\\

STSB                                 & 
85.7                                 & 
{\textbf{79.0}} & 
{\textbf{66.3}} & 
{\textbf{77.3}} & 
66.9                                 & 
62.6                                 & 
{\textbf{87.3}} & 
{\textbf{90.7}} & 
84.7                                 & 
{\textbf{77.8}}  
\\
 
QQP                                  & 
82.1                                 & 
{\textbf{78.0}} & 
{\textbf{65.4}} & 
{\textbf{80.1}} &
64.6                                 & 
56.3                                 & 
{\textbf{87.0}} & 
90.5                                 & 
83.7                                 & 
\textbf{76.4}                        
\\ \midrule
 
SQuAD                                & 
87.5                                 & 
74.0                                 & 
{\textbf{66.3}} & 
71.8                                 & 
51.7                                 & 
6.0                                  & 
{\textbf{87.3}} & 
89.3                                 & 
82.5                                 & 
\textbf{68.5}                        
\\
 
CoNLL03                              & 
73.2                                 & 
64.0                                 & 
63.5                                 & 
60.3                                 & 
51.9                                 & 
0.0                                  & 
71.3                                 & 
16.4                                 & 
84.8                                 & 
\textbf{53.9}                        
\\
 
CoNLL04                              & 
82.1                                 & 
75.0                                 & 
{\textbf{66.3}} & 
{\textbf{76.5}} & 
{\textbf{67.1}} & 
{\textbf{64.3}} & 
{\textbf{87.5}} & 
87.9                                 & 
85.5                                 & 
\textbf{76.9}                        
\\
 
CoNLL05                              & 
87.5                                 & 
65.0                                 & 
64.4                                 & 
69.3                                 & 
{\textbf{68.3}} & 
61.3                                 & 
{\textbf{88.7}} & 
88.4                                 & 
83.8                                 & 
\textbf{75.2}                       
\\
 
CoNLL12                              & 
{\textbf{89.3}} & 
62.0                                 & 
{\textbf{67.3}} & 
63.2                                 & 
{\textbf{67.4}} & 
58.7                                 & 
85.4                                 & 
88.5                                 & 
83.6                                 & 
\textbf{73.9}                        
\\

Ontonotes                            & 
78.6                                 & 
65.0                                 & 
{\textbf{66.3}} & 
56.7                                 & 
54.1                                 & 
59.3                                 & 
82.4                                 & 
84.5                                 & 
{\textbf{86.1}} & 
\textbf{70.3}            
\\ 
\hdashline
\multicolumn{11}{l}{\textit{Best transfer results of the vanilla prompt transfer}} \\
 ORACLE&96.4 &79.0 &67.3 &80.9 &70.4 &64.5 &88.7 &91.0 &86.1 &\cellcolor[HTML]{FFFFFF}\textbf{80.4} \\
 \hdashline
\multicolumn{11}{l}{\textit{Best of Top-3 transfer results predicted by ``E$_{avg}$'' metric in SPoT~\cite{vu2021spot}. }} \\
-w ``E$_{avg}$'' &96.4 &71.0 &67.3 &80.9 &66.8 &61.3 &88.5 &91.0 &86.1 &\cellcolor[HTML]{FFFFFF}\textbf{78.8} \\
\midrule
\multicolumn{11}{c}{\textbf{\textit{(b) Transfer with Our \our approach.}} \zqh{P-Tuning-v2 is used for prompt-tuning in PoT.}}  \\
\midrule
 
BoolQ                                & 
{\textbf{89.3}} &
75.0                                 &
64.4                                 &
{\textbf{76.9}} & 
{\textbf{68.7}} & 
63.7                                 &
{\textbf{88.0}} & 
{90.6}          &
{\textbf{86.3}} &
{\textbf{78.1}}$\boldsymbol{_{2.6}}$                      
\\

CB                                   &
87.5                                 & 
76.0                                 & 
{\textbf{67.3}} & 
71.1                                 & 
{\textbf{69.3}} & 
{\textbf{64.6}} & 
{\textbf{88.7}} & 
{\textbf{90.6}} &
85.5                                 &
{\textbf{77.8}}$\boldsymbol{_{0.8}}$                        
\\

RTE                                  &
{\textbf{92.9}} &
73.0                                 &
{\textbf{66.6}} & 
76.2                                 & 
{\textbf{68.7}} &
{\textbf{64.2}} &
{\textbf{87.3}} & 
{\textbf{90.7}} &
{\textbf{85.8}} & 
{\textbf{78.4}}$\boldsymbol{_{1.4}}$                         
\\

WIC                                  & 
{\textbf{91.1}} &
{\textbf{78.0}} & 
{\textbf{66.3}} &
{\textbf{77.6}} &
66.9                                 &
{\textbf{64.3}} &
{\textbf{87.5}} &
90.5                                 &
{\textbf{86.0}} &
\cellcolor[HTML]{FFFFFF}{\textbf{78.7}}$\boldsymbol{_{3.2}}$  
\\

WSC                                  &
{\textbf{89.3}} &
75.0                                 &
64.4                                 &
76.2                                 &
{\textbf{69.5}} &
63.8                                 &
{\textbf{89.7}} & 
{\textbf{90.7}} &
84.7                                 & 
{\textbf{78.1}}$\boldsymbol{_{0.3}}$                        
\\

COPA                                 &
{\textbf{91.1}} & 
76.0                                 &
64.4                                 &
{\textbf{76.9}} &
{\textbf{69.4}} & 
62.4                                 &
86.8                                 & 
{\textbf{90.6}} & 
84.8                                 & 
{\textbf{78.0}}$\boldsymbol{_{0.8}}$        
\\

MultiRc                              &
{\textbf{91.1}} &
76.0                                 & 
63.5                                 &
{\textbf{77.3}} &
{\textbf{68.0}} &
62.3                                 &
{\textbf{89.0}} & 
{90.6}          & 
{\textbf{85.6}} &
{\textbf{78.2}}$\boldsymbol{_{1.1}}$  
\\

ReCoRD                               &
87.5                                 &
76.0                                 &
{\textbf{66.3}} &
{\textbf{77.3}} &
{\textbf{68.5}} &
62.4                                 & 
{\textbf{87.5}} &
{\textbf{90.7}} &
84.9                                 & 
{\textbf{77.9}}$\boldsymbol{_{15.9}}$                          
\\ \hline

MNLI                                 &
{\textbf{92.9}} & 
{\textbf{77.0}} &
{\textbf{67.3}} & 
{\textbf{78.0}} & 
{\textbf{68.8}} &
{\textbf{66.3}} &
{\textbf{88.5}} & 
{\textbf{90.6}} &
85.4                                 & \textbf{79.4}$\boldsymbol{_{1.3}}$  
\\

CoLA                                 & 
{\textbf{94.6}} &
{\textbf{78.0}} & 
{\textbf{66.3}} &
75.8                                 &
{\textbf{68.5}} & 
{\textbf{65.3}} &
{\textbf{88.0}} & 
{90.6}          &
84.9                                 &
\cellcolor[HTML]{FFFFFF}{\textbf{79.1}}$\boldsymbol{_{2.1}}$ 
\\

SST2                                 &
{\textbf{92.9}} & 
{\textbf{77.0}} & 
{\textbf{68.3}} & 
{\textbf{76.5}} &
{\textbf{70.1}} &
{\textbf{64.8}} &
{\textbf{88.5}} &
{\textbf{90.7}} &
{\textbf{86.3}} & 
\cellcolor[HTML]{FFFFFF}{\textbf{79.5}}$\boldsymbol{_{2.7}}$ 
\\

QNLI                                 & 
{\textbf{92.9}} &
{\textbf{77.0}} & 
{\textbf{66.3}} &
{\textbf{77.3}} &
{\textbf{70.8}} & 
{63.9}          &
{\textbf{87.5}} & 
{\textbf{90.8}} & 
{\textbf{86.6}} & 
\cellcolor[HTML]{FFFFFF}{\textbf{79.2}}$\boldsymbol{_{1.0}}$  
\\

MRPC                                 &
{\textbf{91.1}} &
75.0                                 &
{\textbf{67.3}} &
{\textbf{76.5}} & 
{\textbf{68.5}} &
{\textbf{64.2}} &
{88.0}          &
{\textbf{90.7}} & 
{\textbf{86.3}} & 
\cellcolor[HTML]{FFFFFF}{\textbf{78.6}}$\boldsymbol{_{0.5}}$  
\\

STSB                                 & 
{\textbf{92.9}} & 
76.0                                 & 
{\textbf{67.3}} & 
75.8                                 & 
{\textbf{69.0}} & 
{64.0}          & 
{\textbf{88.7}} & 
90.5                                 &
85.5                                 & 
\cellcolor[HTML]{FFFFFF}{\textbf{78.9}}$\boldsymbol{_{1.0}}$  
\\

QQP                                  &
{\textbf{94.6}} &
{\textbf{77.0}} &
{\textbf{66.3}} &
76.2                                 & 
{\textbf{69.4}} & 
62.6                                 & 
{\textbf{87.0}} & 
{\textbf{90.7}} &
{86.0}          & 
\cellcolor[HTML]{FFFFFF}{\textbf{78.9}}$\boldsymbol{_{2.5}}$  
\\ \hline

SQuAD                                & 
{\textbf{89.3}} & 
75.0                                 & 
{\textbf{66.3}} &
75.5                                 &
{\textbf{69.3}} &
63.1                                 &
{\textbf{87.3}} &
88.9                                 &
{\textbf{85.7}} &
{\textbf{77.8}}$\boldsymbol{_{9.3}}$                           
\\

CoNLL03                              & 
{\textbf{91.1}} &
72.0                                 & 
{\textbf{68.3}} & 
{\textbf{76.9}} & 
{\textbf{67.4}} &
63.6                                 &
86.5                                 &
{\textbf{90.6}} & 
{\textbf{85.6}} & 
{\textbf{78.0}}$\boldsymbol{_{24.1}}$                
\\

CoNLL04                              &
87.5                                 &
73.0                                 & 
{\textbf{68.3}} & 
75.1                                 &
65.0                                 &
{\textbf{64.1}} &
{\textbf{91.0}} &
{\textbf{90.7}} & 
{\textbf{86.2}} & 
{\textbf{77.9}}$\boldsymbol{_{1.0}}$                 
\\

CoNLL05                              & 
87.5                                 &
{\textbf{79.0}} &
{\textbf{65.4}} & 
{\textbf{77.6}} & 
{\textbf{69.6}} & 
{\textbf{63.7}} & 
{\textbf{87.5}} &
{\textbf{90.8}} & 
84.8                                 & 
{\textbf{78.4}}$\boldsymbol{_{3.2}}$                     
\\

CoNLL12                              & 
87.5                                 & 
76.0                                 & 
{\textbf{66.3}} & 
74.4                                 &
{\textbf{68.5}} &
{\textbf{63.7}} &
{\textbf{87.5}} &
{\textbf{90.8}} & 
85.0                                 & 
{\textbf{77.7}}$\boldsymbol{_{3.8}}$    
\\

Ontonotes                            & 
{\textbf{89.3}} & 
74.0                                 & 
{\textbf{66.3}} &
76.2                                 & 
{\textbf{69.1}} &
{\textbf{64.2}} & 
{\textbf{88.0}} & 
{\textbf{90.8}} &
{\textbf{85.7}} &
{\textbf{78.2}}$\boldsymbol{_{7.9}}$      \\
 \hdashline
 \multicolumn{11}{l}{\textit{Best transfer results of our \our approach}} \\
 ORACLE &94.6 &79.0 &68.3 &78.0 &70.8 &66.3 &91.0 &90.8 &86.6 &\cellcolor[HTML]{FFFFFF}\textbf{80.6}$\boldsymbol{_{0.2}}$ \\
  \hdashline
\multicolumn{11}{l}{\textit{Best of Top-3 transfer results predicted by our metric. }} \\
-w Ours &92.9 &79.0 &68.3 &78.0 &70.8 &66.3 &88.5 &90.8 &86.2 &\cellcolor[HTML]{FFFFFF}\textbf{80.1}$\boldsymbol{_{1.3}}$ \\
\bottomrule       

\end{tabular}
}
\caption{Results (\%) of cross-tasks PoT based on BERT-large model.
In the groups of (a) and (b), each cell denotes the target task performance when transferring the prompt from source task (row) to the associated target task (column).
``AVG.'' denotes the average performance of all target tasks. Notably, positive prompt transfers are in \textbf{bold} and numbers in the subscript indicate relative improvements of our \our against vanilla PoT.}
\label{tab:results_large}
\end{table*}
\begin{table*}[ht]
\centering
\setlength{\tabcolsep}{14pt}
\scalebox{1}{
\renewcommand\arraystretch{1}
\begin{tabular}{lcccccccc}
\toprule
                         &
 \multicolumn{2}{c}{\textbf{BERT-base$_{110M}$}}                                                 &
 \multicolumn{2}{c}{\textbf{BERT-medium$_{42M}$}}                                         &
 \multicolumn{2}{c}{\textbf{BERT-small$_{29M}$}}                                                &
 \multicolumn{2}{c}{\textbf{BERT-tiny$_{4.4M}$}} \\ \cmidrule(lr){2-3} \cmidrule(lr){4-5} \cmidrule(lr){6-7} \cmidrule(lr){8-9}
\multirow{-2}{*}{\textbf{Method}} &
 Vanilla                           &
 \our                           &
 Vanilla                            &
 \multicolumn{1}{c}{\our} &
 Vanilla                            &
 \our                           &
 Vanilla   &
 \our    \\ \midrule \midrule
model-tuning              &
 \multicolumn{2}{c}{74.8}                                                      &
 \multicolumn{2}{c}{71.0}                                                &
 \multicolumn{2}{c}{68.7}                                                      &
 \multicolumn{2}{c}{60.8}      \\
\zqh{Lester et al.~\cite{lester2021power}}              &
 \multicolumn{2}{c}{\zqh{67.5}}                                                      &
 \multicolumn{2}{c}{\zqh{66.5}}                                                &
 \multicolumn{2}{c}{\zqh{62.5}}                                                      &
 \multicolumn{2}{c}{\zqh{58.1}}      \\
\zqh{P-Tuning-v2~\cite{liu2021p}}             &
 \multicolumn{2}{c}{74.6}                                                      &
 \multicolumn{2}{c}{70.5}                                                &
 \multicolumn{2}{c}{67.5}                                                      &
 \multicolumn{2}{c}{59.1}      \\ \midrule
MNLI                     &
 \cellcolor[HTML]{FFFFFF}\textbf{75.6} &
 \cellcolor[HTML]{FFFFFF}\textbf{75.9} &
 \cellcolor[HTML]{FFFFFF}\textbf{71.3} &
 \cellcolor[HTML]{FFFFFF}\textbf{71.7} &
 \cellcolor[HTML]{FFFFFF}\textbf{68.1} &
 \cellcolor[HTML]{FFFFFF}\textbf{69.2} &
 \cellcolor[HTML]{FFFFFF}59.8 &
 \cellcolor[HTML]{FFFFFF}60.2          \\
QNLI                     &
 \cellcolor[HTML]{FFFFFF}\textbf{75.1} &
 \cellcolor[HTML]{FFFFFF}\textbf{75.6} &
 69.2 &
 \cellcolor[HTML]{FFFFFF}\textbf{72.5} &
 67.5                                  &
 \cellcolor[HTML]{FFFFFF}\textbf{69.3} &
 \cellcolor[HTML]{FFFFFF}59.2 &
 \cellcolor[HTML]{FFFFFF}\textbf{60.5} \\
Record                   &
 74.1                                  &
 \cellcolor[HTML]{FFFFFF}\textbf{75.2} &
 67.2                                 &
 \cellcolor[HTML]{FFFFFF}\textbf{71.3} &
 62.1                                  &
 \cellcolor[HTML]{FFFFFF}\textbf{68.7} &
 45.4                         &
 \cellcolor[HTML]{FFFFFF}\textbf{60.5} \\
SQuAD                    &
 73.8                                  &
 \cellcolor[HTML]{FFFFFF}\textbf{75.1} &
 69.3                                 &
 \cellcolor[HTML]{FFFFFF}\textbf{71.6} &
 67.2                                  &
 \cellcolor[HTML]{FFFFFF}\textbf{68.4} &
 58.7                         &
 \cellcolor[HTML]{FFFFFF}\textbf{61.2} \\
CoNLL03                  &
 68.1                                  &
 \cellcolor[HTML]{FFFFFF}\textbf{75.7} &
 67.6                                  &
 \cellcolor[HTML]{FFFFFF}\textbf{71.2} &
 65.9                                  &
 \cellcolor[HTML]{FFFFFF}\textbf{69.2} &
 58.0                         &
 \cellcolor[HTML]{FFFFFF}\textbf{60.7} \\
Ontonotes                &
 71.2                                  &
 \cellcolor[HTML]{FFFFFF}\textbf{75.3} &
 68.4                                  &
 \cellcolor[HTML]{FFFFFF}\textbf{71.6} &
 65.0                                  &
 \cellcolor[HTML]{FFFFFF}\textbf{68.6} &
 58.2                         &
 \cellcolor[HTML]{FFFFFF}\textbf{60.0} \\
CoNLL05                  &
 72.8                                  &
 \cellcolor[HTML]{FFFFFF}\textbf{75.9} &
 69.5                                 &
 \cellcolor[HTML]{FFFFFF}\textbf{71.8} &
 66.0                                  &
 \cellcolor[HTML]{FFFFFF}\textbf{69.6} &
 57.8                         &
 \cellcolor[HTML]{FFFFFF}\textbf{60.2} \\
CoNLL12                  &
 73.8                                  &
 \cellcolor[HTML]{FFFFFF}\textbf{75.2} &
 69.9                                 &
 \cellcolor[HTML]{FFFFFF}\textbf{71.7} &
 66.5                                  &
 \cellcolor[HTML]{FFFFFF}\textbf{68.6} &
 58.4                         &
 \cellcolor[HTML]{FFFFFF}\textbf{59.9} \\
SST2                     &
 73.8                                  &
 \cellcolor[HTML]{FFFFFF}\textbf{76.3} &
 70.0                               &
 \cellcolor[HTML]{FFFFFF}\textbf{71.9} &
 67.1                                  &
 \cellcolor[HTML]{FFFFFF}\textbf{69.1} &
 58.3                         &
 \cellcolor[HTML]{FFFFFF}\textbf{60.0}
\\ \bottomrule
\end{tabular}
}
\caption{Average performance (\%) of all target datasets (full results can be seen in Appendix). Note that the column ``Vanilla'' denotes the results of vanilla PoT. Positive prompt transfers are in \textbf{bold}. Numbers in the subscript show the corresponding model parameters of these backbone models.}
\label{tab:main2}
\end{table*}

\section{Experiments}
\label{sec_experiment}
\subsection{Tasks and Datasets} \label{task_and_data}
To investigate the effectiveness of our proposed methods, we conduct comparative experiments on a total of 189 diverse combinations of 21 source datasets and 9 target datasets, which comprise part of tasks from General Language Understanding Evaluation~(GLUE)~\cite{wang2018glue}, SuperGLUE~\cite{wang2019superglue} and other NLU benchmarks. Specifically, GLUE is one of the most popular NLP benchmarks, which consists of several NLU tasks that can be  covering sentiment analysis, question answering and natural language inference. We use most of GLUE datasets as source datasets, including MNLI, CoLA, SST2, QNLI, MRPC, STSB and QQP. Moreover, as a stickier benchmark, SuperGLUE provides a new set of more difficult language understanding tasks, which additionally include word sense disambiguation and multi-choice question answering. All datasets of SuperGLUE are collected into source datasets, \textit{i.e.} BoolQ, CB, RTE, WIC, WSC, COPA, Multirc and Record. Furthermore, to enrich the types of source datasets, some other NLU benchmarks are also added, including extractive QA~(SQuAD 2.0~\cite{rajpurkar2016squad}), named entity recognition~(NER) (CoNLL03~\cite{sang2003introduction}, CoNLL04~\cite{carreras-marquez-2004-introduction} and Ontonotes~\cite{hovy2006ontonotes}) and semantic role labeling~(SRL) (CoNLL05~\cite{carreras2005introduction} and CoNLL12~\cite{pradhan-etal-2012-conll}). Following the prior work~\cite{vu2021spot}, we use low-resource datasets as target tasks to simulate a realistic scenario, which covers SuperGLUE~(CB, COPA, WSC, RTE, WIC), GLUE~(CoLA, MRPC, STSB) and other task~(CoNLL04). 

For most datasets, we use accuracy (Acc.) as their evaluation metric in the experiments, except the Matthews correlation (Mcc.) for CoLA, F1 score (F1) for Multirc, Record, SQuAD, CoNLL03, CoNLL04, CoNLL05, CoNLL12 and Ontonotes, the combined score of Pearson-Spearman correlations (Pcor/Scor.) for STS-B. Notably, due to restricted test set access for GLUE and SuperGLUE, we follow Vu~\textit{et~al.}~\cite{vu2021spot} to train with a ﬁxed number of steps\footnote{{\zqh The training steps depend on the scale of training corpus, batch size and training epochs. We have shown the detailed information in Table~\ref{tab:dataset_details}.}} and report results on the validation set associated with each dataset. Table~\ref{tab:dataset_details} shows the {\zqh statistic and training} details of all 21 tasks used in our study. The tasks are classified into three groups: SuperGLUE, GLUE and others.

\subsection{Implementation Details} 
\label{implementation}
We adopt our methods to fine-tune 5 different scales of PLMs, \textit{i.e.}, BERT-large~(340M), BERT-base~(110M), BERT-medium~(42M), BERT-small~(29M) and BERT-tiny~(4.4M). The cutting-edge method P-Tuning-v2~\cite{liu2021p} is introduced to perform basic prompt-tuning. We use AdamW optimizer to tune our models with 8 NVIDIA A100 GPUs. For each dataset, we perform a grid search for the learning rate on \{5e-3, 7e-3, 1e-2\}, the batch size on \{16, 32, 64\}, and epoch on \{20, 40, 60, 80, 100\}. The dropout rate is set to 0.1 and the input sequence length is 128. For the prompt length, we follow prior work~\cite{vu2021spot} and set it to a fixed number of 20 tokens in all cases\footnote{{\zqh As stated in the P-Tuning-v2~\cite{liu2021p}, the different tasks achieve their best performance with different prompt lengths. However, in the vanilla PoT scenarios, the prompt lengths across different tasks should be equal, so that the prompt embeddings can be reused between different tasks. Hence, we follow the prior works~\cite{vu2021spot,su2021transferability} and set a fixed prompt length for all tasks, which may cause the performance discrepancy between the original P-Tuning-v2 paper~\cite{liu2021p} and our re-implementations.}}, {\zqh \textit{i.e.}, continuous prompts with 20 token lengths are added in every layer as prefix tokens} (The detailed hyper-parameters are also listed in Table~\ref{tab:dataset_details}). For references, we compare our \our approach with full-parameter model-tuning, regular prompt-tuning {\zqh methods (\textit{i.e.}, the vanilla prompt-tuning method, Lester~et~al.~\cite{lester2021power}, and a more powerful prompt-tuning method, P-Tuning-v2~\cite{liu2021p})} without any transfer, and vanilla PoT methods. {\zqh Specifically, since the SPoT~\cite{vu2021spot} is the first and most representative PoT method, we use it as the main baseline method. Moreover, we also compare our method with the more cutting-edge PoT methods, \textit{i.e.}, ATTEMPT~\cite{asai2022attempt} and SEMoE~\cite{peng2022model}.}
It is noteworthy that most PLMs are fixed during training in all experiments, except those with model-tuning. 
Additionally, note that we evaluate the performance in the full-data scenario for all experiments.

\subsection{Main Results} \label{main_result}
\paragraph{\textbf{Prompt-tuning via \our approach consistently outperforms model-tuning}} Table~\ref{tab:results_large} shows comparative results on 9 target tasks of BERT-large with vanilla PoT and \our respectively. 
Firstly, it can be found that P-Tuning-v2 can achieve better results than model-tuning on some target datasets (\textit{e.g.}, COPA and CoLA), but its average performance of all datasets is still sub-optimal (77.5\% \textit{v.s.} 78.5\%), showing the limitation of prompt-tuning. 
Secondly, in the group (a) of vanilla PoT, we can observe that transferring soft prompt from a similar source task can indeed benefit the target performance, \textit{e.g.,} MNLI (77.5\% $\rightarrow$ 78.1\%) and QNLI (77.5\% $\rightarrow$ 78.2\%), which confirms the significance of PoT. Lastly, \textit{with our \our, the prompt transfer can be more effective and prompt-tuning can achieve comparable and even better performance than model-tuning} (\textit{e.g.}, MNLI: 79.4\% \textit{v.s.} 78.5\%).

\paragraph{\textbf{Knowledge Distillation helps bridge the gap between different types of tasks}} 
Although compared with the basic prompt-tuning (P-Tuning-v2) on the NLI tasks (many of target datasets belong to NLI task), the vanilla PoT can achieve performance improvement, PoT on the other dissimilar tasks (\textit{e.g.}, NER and SRL) is usually negative and even leads to worse performance.
This indicates that the large gap between source and target tasks would hinder knowledge transfer, confirming our statement.
On the contrary, \our can consistently boost the performance of prompt-tuning on all types of source tasks. More specifically, \our significantly surpasses the vanilla PoT in the scenario where the source task is dissimilar to the target task. For instance, when using CoNLL03 (NER) and Record (QA) as source tasks, the relative improvements of \our are up to 24.1\% and 15.9\%, respectively. \textit{We state that this is owing to the knowledge distillation technique, which can effectively bridge the gap between different types of tasks and alleviate the problem of knowledge forgetting.}

\begin{table}[t]
    \centering
    \setlength{\tabcolsep}{15pt}
    \scalebox{1}{
    \begin{tabular}{lcc}
    \toprule
    \multicolumn{1}{c}{\textbf{Method}}                               & \textbf{Performance}            & \textbf{Gains ($\uparrow$)}           \\  \midrule \midrule
    \multicolumn{1}{l}{\zqh{P-Tuning-v2~\cite{liu2021p}}}                          & 74.6                 & -                 \\ \hdashline
    \multicolumn{2}{l}{\textit{\textbf{Single-task PoT Settings}}}  \\
    \multicolumn{1}{l}{\quad vanilla PoT~\cite{vu2021spot} (avg.)}                          & 75.0                 & +0.4                \\ 
    \multicolumn{1}{l}{\quad \our (avg.)}                          & 75.8                 & +1.2                \\ 
    \midrule
    \multicolumn{2}{l}{\textit{\textbf{Multi-task PoT Settings}}}   \\
    \multicolumn{1}{l}{\quad ATTEMPT~\cite{asai2022attempt}} & 75.8 & +1.2 \\
    \quad SESoM~\cite{peng2022model}                     & 75.1                 & +0.5                \\ \hdashline
    \quad \our-Early                  & \textbf{76.5}                 & \textbf{+1.9}                \\ 
    \quad \our-Late                  & 76.3                 & +1.7                \\ 
    \bottomrule
    \end{tabular}
    }
    \caption{Comparison of single-task PoT and multi-task PoT. MNLI, QNLI and QQP are used as source tasks and average scores (\%) of 9 target tasks are reported. ``avg.'' denotes that we average the PoT results of different source-target pairs.}
    \label{tab:multi_task}
\end{table}

\paragraph{\textbf{Better correlation between predicted prompt transferability and PoT performance}} To verify the effectiveness of prompt transferability metric, we follow SPoT~\cite{vu2021spot} and use the metric to predict the Top-3 source tasks for each target task and report the best of Top-3 transfer results in Table~\ref{tab:results_large} as well. For reference, the oracle best transfer results are also reported. It can be seen, compared to the ``E$_{avg}$'' metric in SPoT, our metric can predict the source tasks more accurately and achieve better PoT performance based on \our. That is, \textit{there is a better correlation between predicted prompt transferability and PoT performance when using our metric in \our framework.}

\paragraph{\textbf{\our shows generality and applicability on various scales of PLMs}} We also conduct experiments on other sizes of PLMs and show the average results in Table~\ref{tab:main2}. 
As seen, our \our consistently boosts the effectiveness of PoT and improves the performance of prompt-tuning over the model-tuning approach across all model sizes. {\zqh Notably, in the extremely small BERT-tiny setting, it is difficult for both P-Tuning-v2 and our method to achieve comparable performance as model-tuning. This phenomenon is similar to the findings of Lester~et~al.~\cite{lester2021power} that claim \textit{prompt-tuning becomes more competitive with scale}\footnote{{\zqh One of our conjectures is that prompt-tuning may fall short in transferring the limited pretrained knowledge of smaller PLMs to downstream tasks. We will perform more in-depth analyses for it in our future work.}}. Nevertheless, our \our still has the great potential to help prompt-tuning achieve better performance against model-tuning.} \textit{In general, these results can demonstrate the generality and applicability of our \our approach on various scales of PLMs.}

\paragraph{\textbf{\our can also work well in multi-task scenarios}} In addition to the above single source-target PoT scenarios, we also conduct experiments to verify the effectiveness of \our in the multiple-task scenarios. Specifically, taking the ``MNLI, QNLI, QQP" as source tasks, we report the PoT results of BERT-base on 9 target tasks in Table~\ref{tab:multi_task}. For references, we also report the contrastive results of some cutting-edge multi-task counterparts, \textit{i.e.,} ATTEMPT~\cite{asai2022attempt} and SESoM~\cite{peng2022model}.
As seen, multi-task PoT generally outperforms single-task PoT, and with this strategy, our PANDA can achieve even better results, demonstrating the complementarity of PANDA and multi-task learning. More encouragingly, compared to the powerful counterparts, \our with both fusion strategies can consistently achieve much better performance. \textit{These results demonstrate the effectiveness and superiority of \our in multi-task scenarios.}

\subsection{Ablation Study and Analysis}

\begin{table}[t]
    \centering
    \makeatletter\def\@captype{table}
    \setlength{\tabcolsep}{13pt}
    \scalebox{1}{
    \begin{tabular}{lcc}
    \toprule
    \multicolumn{1}{c}{\textbf{Method}}                               & \textbf{BERT-large}            & \textbf{BERT-small}           \\  \midrule \midrule
    \multicolumn{1}{c}{\zqh{P-Tuning-v2~\cite{liu2021p}}}                          & 77.5                 & 67.5                 \\
    \multicolumn{1}{c}{vanilla PoT~\cite{vu2021spot}}                          & 78.1                 & 68.1                 \\ \midrule
    \textbf{\our}                                &  & \\
    \multicolumn{1}{l}{\quad -w random prompt} & 77.0 & 66.8 \\
    \quad-w source prompt                     & 78.8                 & 68.9                 \\
    \quad -w target-like prompt                  & \textbf{79.4}                 & \textbf{69.2}                \\ 
    \bottomrule
    \end{tabular}
    }
    \centering
    \caption{Comparisons of different teacher prompts used in \our approach. MNLI is used as the source task and average scores (\%) of 9 target tasks are reported.}
    \label{tab:tearcher}
\end{table}

\begin{figure*}[t]
    \begin{minipage}[t]{0.33\textwidth}
    \centering
    \includegraphics[width=\textwidth]{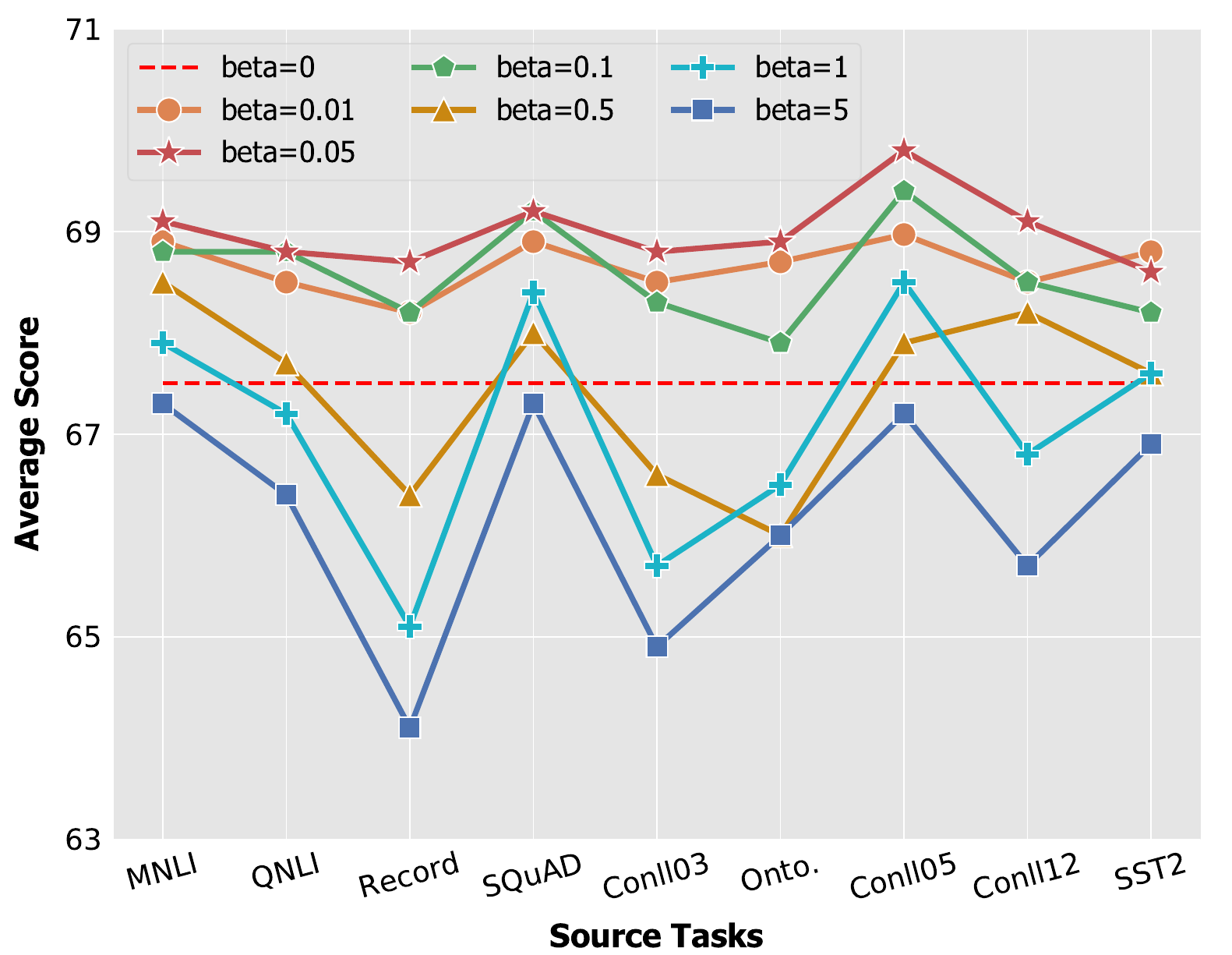}
    \end{minipage}%
    \hfill
    \hfill
    \begin{minipage}[t]{0.33\textwidth}
    \centering
    \includegraphics[width=\textwidth]{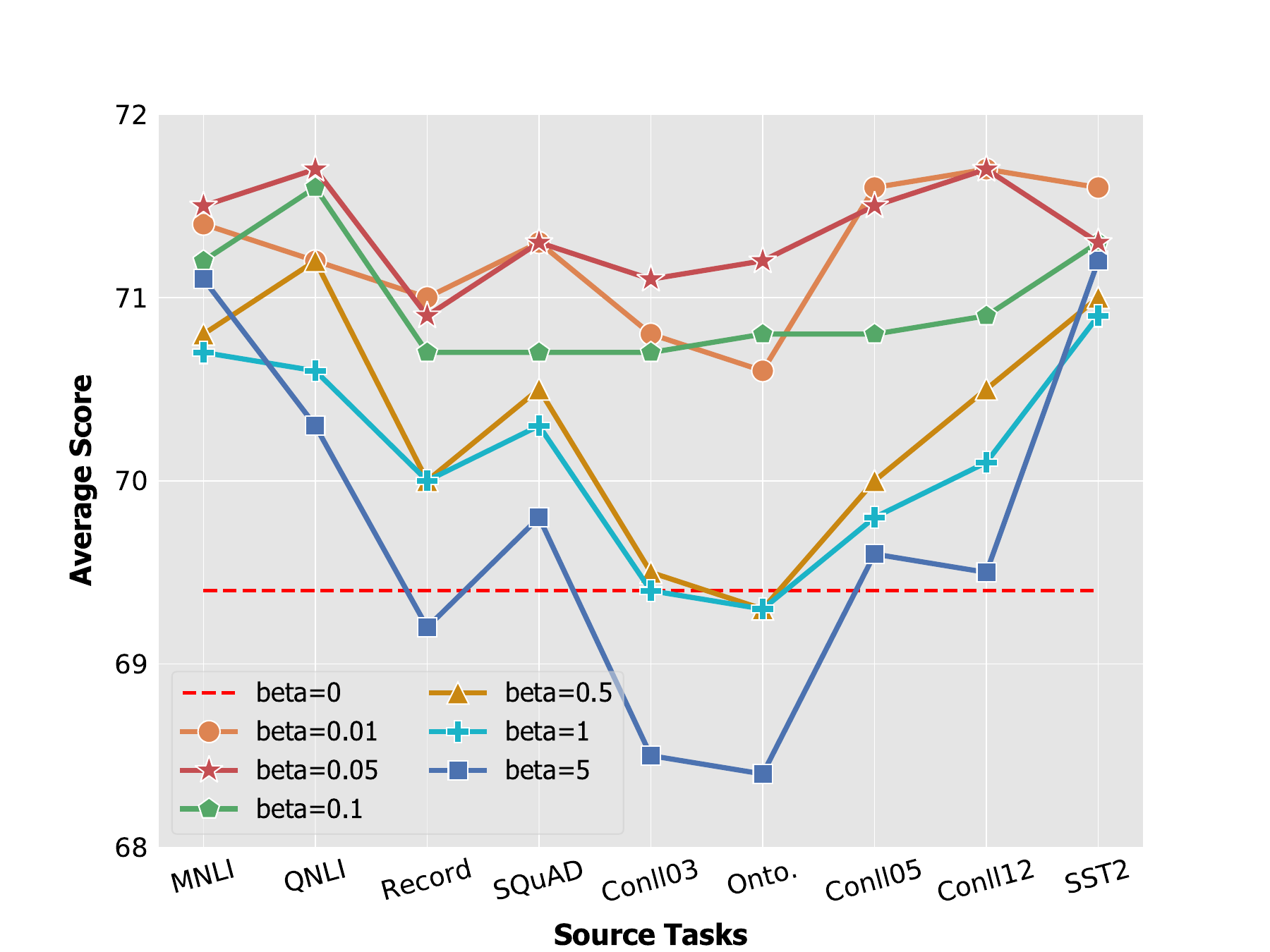}
    \end{minipage}%
    \hfill
    \hfill
    \begin{minipage}[t]{0.33\textwidth}
    \centering
    \includegraphics[width=\textwidth]{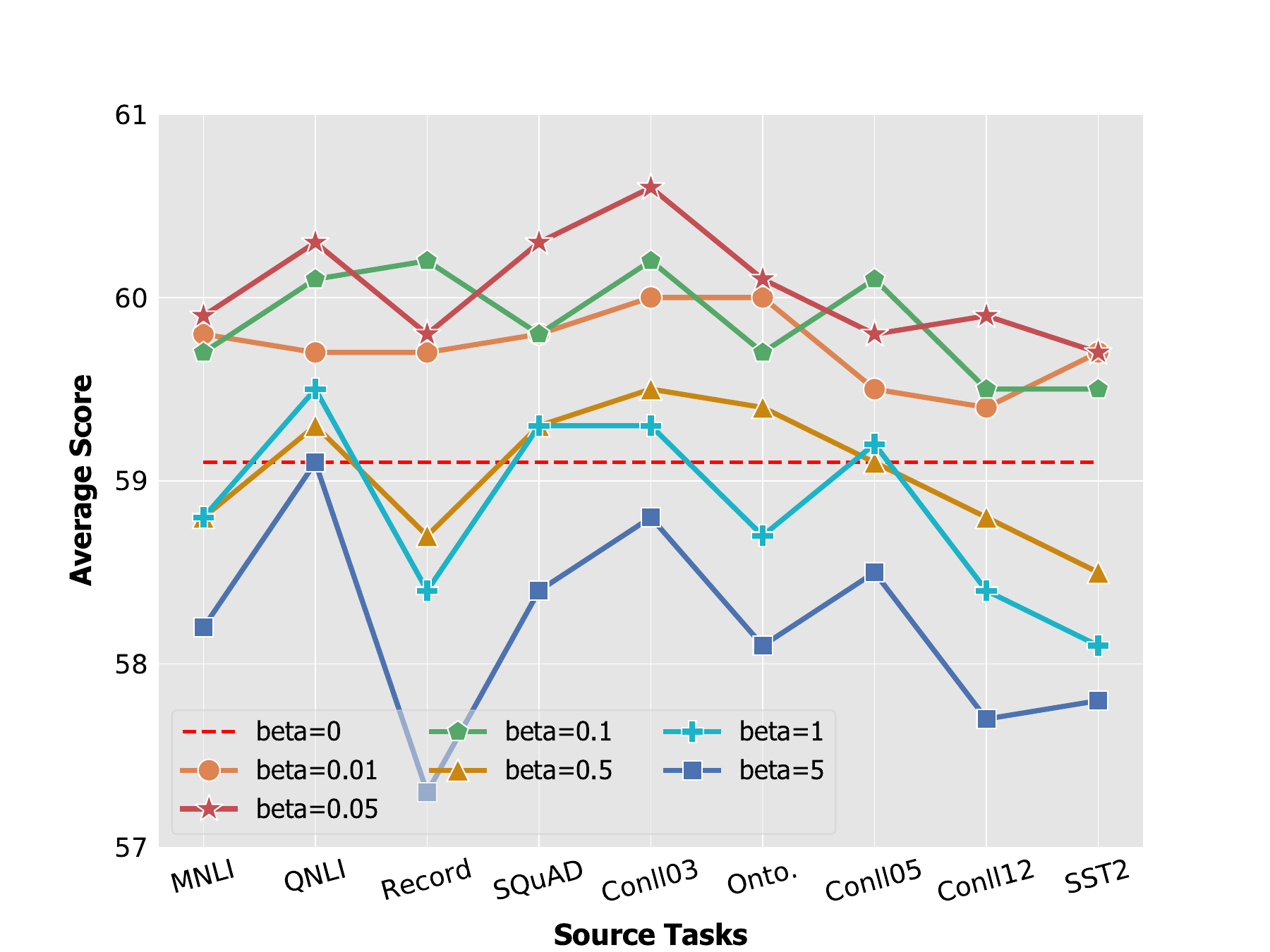} 
    \end{minipage}
    \caption{\textbf{Left}: results of different $\lambda$ on BERT-small. \textbf{Medium}: results of different $\lambda$ on BERT-medium. \textbf{Right}: results of different $\lambda$ on BERT-tiny.
    }
    \label{fig:appendix_c}
\end{figure*}

\paragraph{\textbf{Influence of different teacher prompts}} \label{teacher_ablation}
As mentioned in Section~\ref{sec_panda}, we use the source prompt as the initial teacher prompt and then train it on the target task with fewer iterations to obtain the target-like prompt.
To investigate whether the teacher prompt provides useful knowledge to the student prompt, we conduct comparative experiments using different teacher prompts, \textit{i.e.}, ``source prompt'', ``target-like prompt'' and ``random prompt'' (the randomly initialized prompt), and Table~\ref{tab:tearcher} lists the results. 
When using ``random prompt'' as a teacher prompt, the average performance is much worse than the others, because the randomly initialized prompt can not provide useful knowledge and even hinders the learning of student prompt.
On the contrary, the other two prompts containing more linguistic knowledge perform much better, proving the effectiveness of KD strategy. Among these prompts, the target-like one achieves the best performance, as it not only preserves the general knowledge from the source task but also bridges the discrepancy between source and target tasks.

\begin{table}[t]
    \centering
    \makeatletter\def\@captype{table}
    \setlength{\tabcolsep}{13pt}
    \scalebox{1}{
    \begin{tabular}{lcc}
    \toprule
    \multicolumn{1}{c}{\textbf{Method}}                              & \textbf{BERT-medium}            & \textbf{BERT-tiny}           \\  \midrule \midrule
    \multicolumn{1}{c}{\zqh{P-Tuning-v2~\cite{liu2021p}}}                          & 70.5                & 59.1               \\
    \multicolumn{1}{c}{vanilla PoT~\cite{vu2021spot}}                          & 69.2             & 59.2              \\ \midrule
    \textbf{\our}                                &  & \\
    \quad -w constant (ones)   & 72.0           &60.1      \\
    \quad-w E$_{avg}$ metric                  & 71.7               &   60.1             \\
    \quad -w ON metric                      & 71.8               &   60.2              \\ 
    \quad -w Our metric      & \textbf{72.5}                & \textbf{60.5}           \\ 
    \bottomrule
    \end{tabular}
    }
    \caption{Comparisons of different metric used to control the knowledge transfer. We use the QNLI as the source task and report the average scores (\%) of 9 target tasks.}
    \label{tab:beta}
\end{table}

\paragraph{\textbf{Whether \our benefits from our metric}} \label{beta_ablation}
 As shown in Equation~\ref{eq_beta}, our metric is used to control the knowledge transfer for each source-target pair. To analyze its effectiveness, we replace our metric with ``constant factor (one) '', and other prior metrics, \textit{i.e.}, ``E$_{avg}$'' and ``ON''. Results of different methods are presented in Table~\ref{tab:beta}. 
Compared with the vanilla PoT, all variants of \our achieve better performance on both PLMs, continuing to confirm the effectiveness of our approach. Additionally, it can be seen that our metric offers the most performance improvements over the plain \our, \textit{i.e.}, \our with constant factor (one), which proves that our metric indeed facilitates the adaptive knowledge transfer and thus boosts the performance of \our.

\begin{figure*}[t]
    \begin{minipage}[t]{0.23\textwidth}
    \centering
    \includegraphics[width=\textwidth]{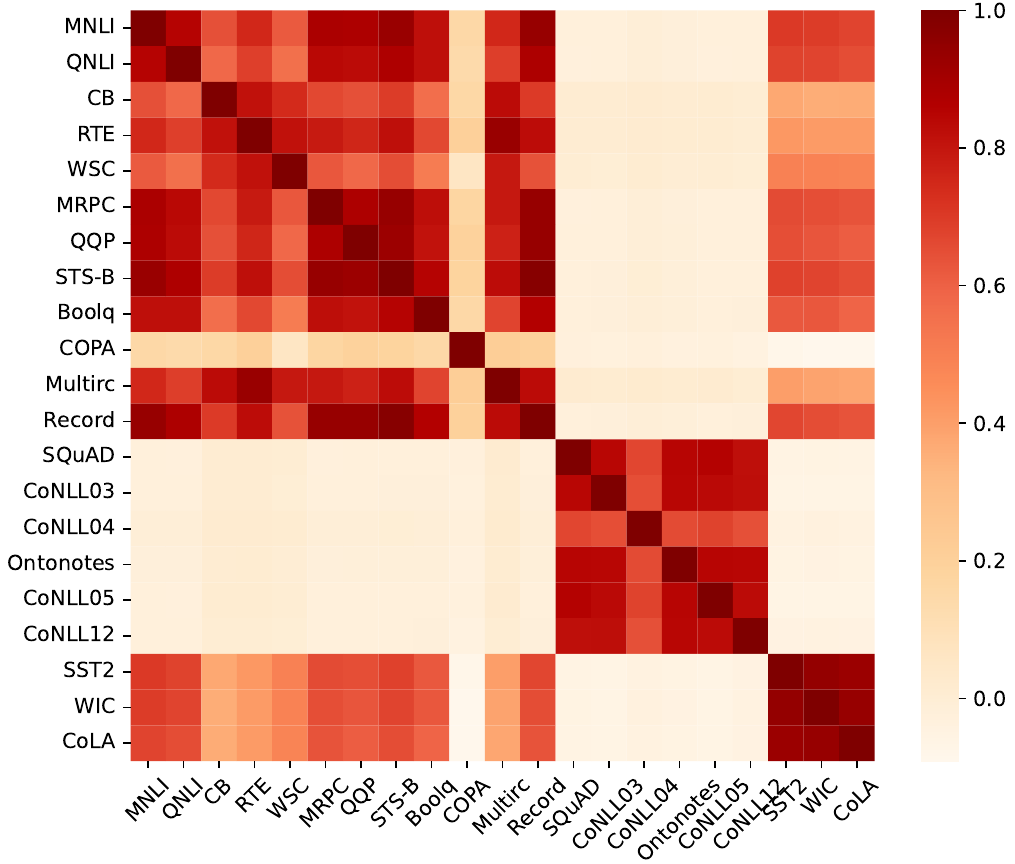}
    \end{minipage}%
    \hfill
    \hfill
    \begin{minipage}[t]{0.23\textwidth}
    \centering
    \includegraphics[width=\textwidth]{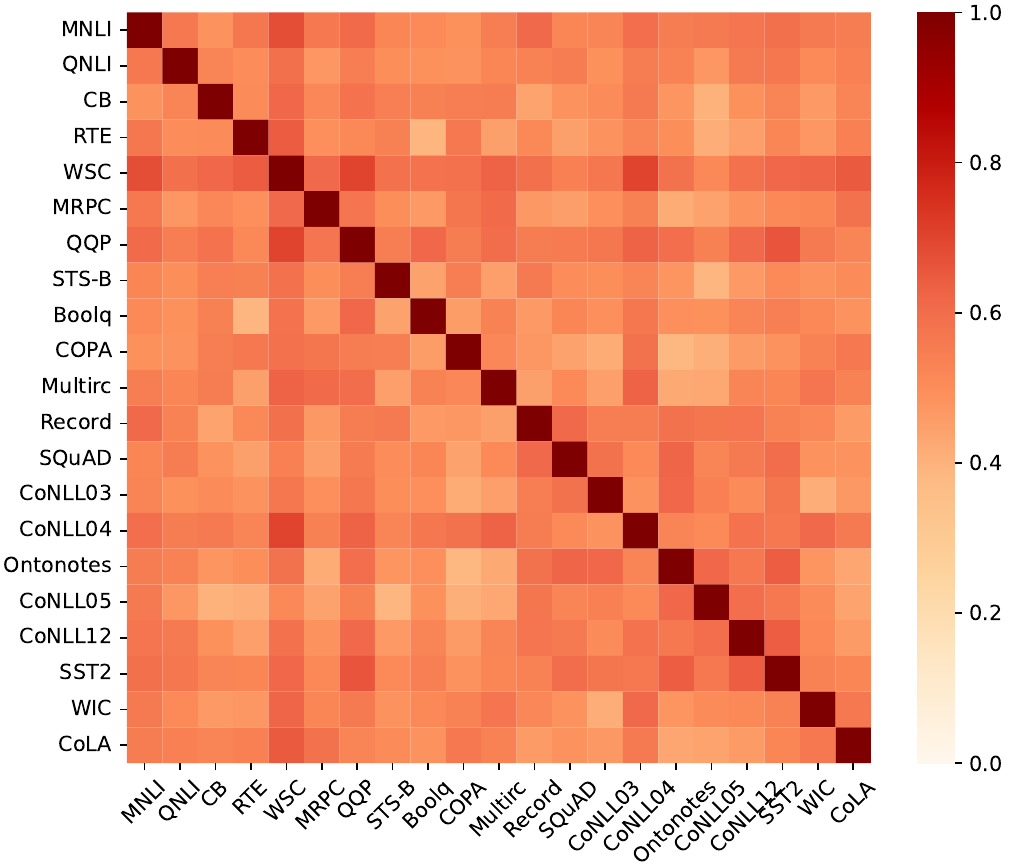} 
    \end{minipage}
    \hfill
    \hfill
    \begin{minipage}[t]{0.228\textwidth}
    \centering
    \includegraphics[width=\textwidth]{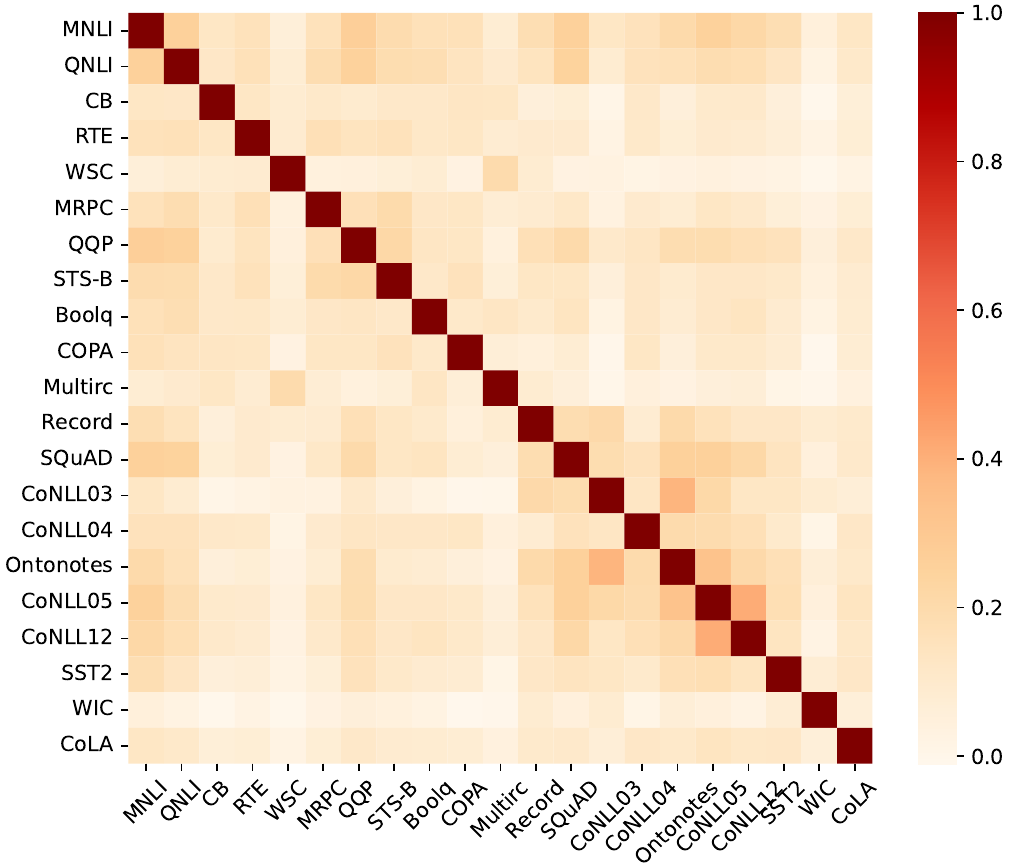} 
    \end{minipage}
    \hfill
    \hfill
    \begin{minipage}[t]{0.26\textwidth}
    \centering
    \includegraphics[width=\textwidth]{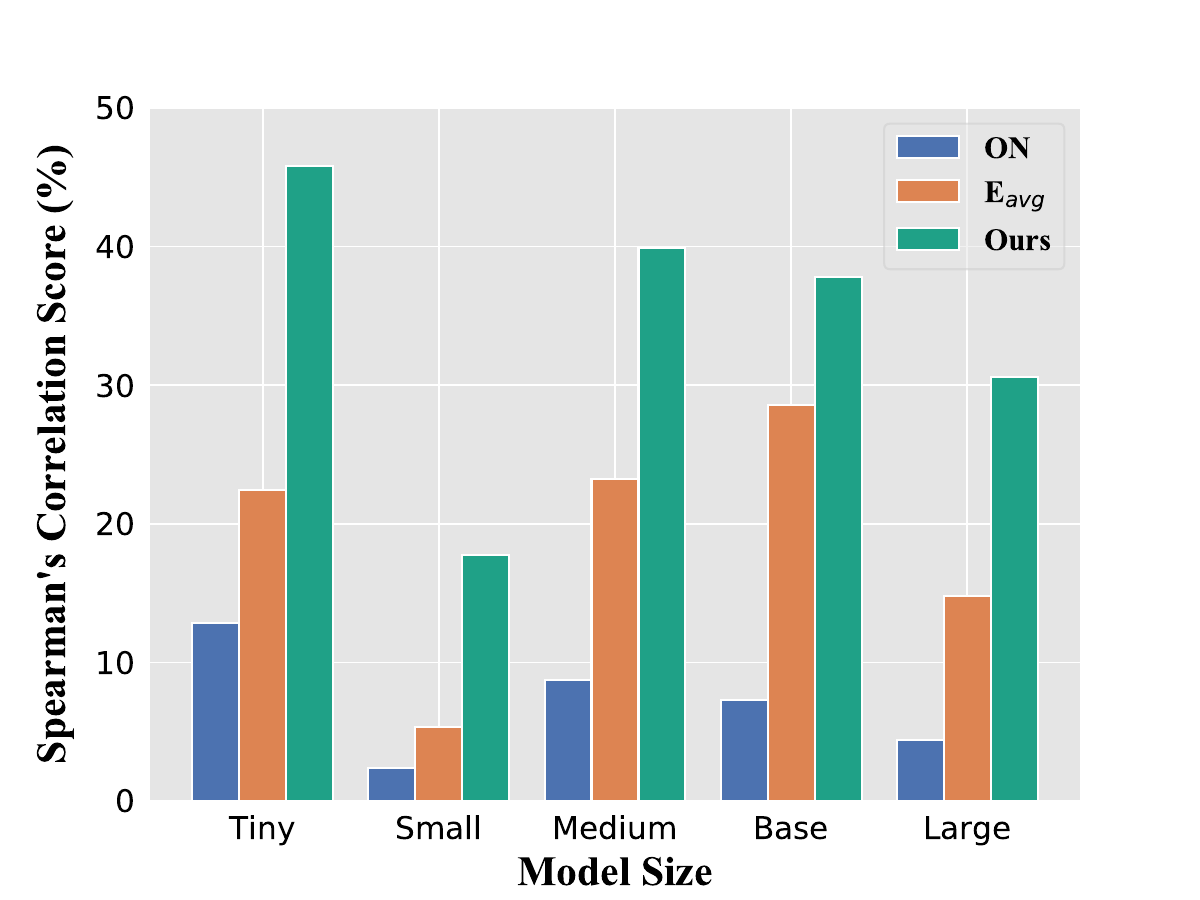} 
    \end{minipage}%
    
    \caption{From \textbf{Left} to \textbf{Right}: 1) a heatmap of our predicted prompt transferability across all 21 tasks; 2): results predicted by the method ``ON'' in prior work~\cite{su2021transferability}; 3): results predicted by the method ``E$_{avg}$'' in the SPoT~\cite{vu2021spot}; 4): Spearman's correlation scores of these metrics with cross-task (21 source tasks $\rightarrow$ 9 target tasks) prompt transfer performance across all model sizes. Note that BERT-large is used in the left three sub-figures, and the results of more models are shown in Appendix.
    }
    \label{fig:heatmap}
\end{figure*}

\paragraph{\textbf{Analysis of different knowledge distillation losses}}
\label{ablation_loss}
{\zqh The core idea of knowledge distillation is to introduce a constraint $\mathcal{L}_{kd}(p, q)$ to encourage the student to match the teacher. It is the cross-entropy (CE) between the teacher and student predictive distributions \{$q, p$\}, both scaled by a temperature hyperparameter $\tau > 0$. As stated in Stanton~et~al.~\cite{stanton2021does}, if $\tau = 1$ then $\mathcal{L}_{kd}$ is similarly equivalent to the Kullback–Leibler divergence (KL). As $\tau \rightarrow +\infty$, $\nabla_{\vec p} \mathcal{L}_{kd}(\vec p, \vec q) \approx \vec q - \vec p$, and thus in the limit $\nabla_{\vec q} \mathcal{L}_{kd}$ is approximately equivalent to $\nabla_{\vec p} || \vec q - \vec p ||^2_2 / 2$, \textit{i.e.}, Mean Squared Error (MSE). As the mainstream distillation loss functions, KL, CE and MSE, have been widely-used in many previous works~\cite{hinton2015distilling, stanton2021does,kim2021comparing}.}
To investigate the effectiveness of different knowledge distillation losses in our \our, we use the above loss functions\footnote{{\zqh Notably, our main contribution is to reveal that prompt transfer benefits from knowledge distillation, but not explore more powerful distillation loss functions. Thus, we use these typical loss functions for simplicity and universality.}} as $\mathcal{L}_{kd}$ in \our and conduct comparative experiments using four source tasks on two PLMs of different scales. Table~\ref{tab:appendix_loss} shows the detailed results. It can be found the different between multiple loss functions is not large, especially the difference between KL and CE. Moreover, we can observe that MSE performs best on most settings, thus we use it as the knowledge distillation loss in \our.

\paragraph{\textbf{Influence of different weighting factor $\lambda$}}
\label{ablation_factor}
The weighting factor $\lambda$ in Equation~\ref{eq_beta} for the knowledge distillation auxiliary task is an important parameter. In the subsection, we evaluate the performance of the proposed \our approach with different $\lambda$ on several pairs of cross-task prompt transfer and show the results in Figure~\ref{fig:appendix_c}.

The weighting factor $\lambda$ affects the knowledge transfer between source and target prompts. On the one hand, setting a too small $\lambda$, \textit{i.e.}, $\lambda$ = 0.01, would make the student prompt ignore the ``dark knowledge'' from teach prompt and thus hinder the knowledge transfer. On the other hand, larger $\lambda$ would prevent the student prompt from learning good knowledge from the target task, making the student prompt perform worse. More importantly, it can be found that a large $\lambda$ is more detrimental than a smaller one. In the case of $\lambda$ = 5, the average performance of all pairs of cross-task prompt transfer is much worse than the case of small $\lambda$, \textit{e.g.}, 0.01 or 0.05. Recall that the case of $\lambda$ = 0.05 performs best among all settings, thus leaving as the default setting.

\begin{table}[t]
    \centering
    \setlength{\tabcolsep}{7pt}
    \scalebox{1}{
    \begin{tabular}{lcccccc}
    \toprule
    & \multicolumn{3}{c}{\textbf{BERT-small}} & \multicolumn{3}{c}{\textbf{BERT-tiny}}           \\ \cmidrule(lr){2-4} \cmidrule(lr){5-7}
    \multirow{-2}{*}{\textbf{Source task}}                     & CE    & KL    & MSE            & CE    & KL             & MSE            \\  \midrule \midrule
    MNLI    & 68.56 & 68.62 & \textbf{69.21} & 59.84 & 59.88          & \textbf{60.2}  \\
    QNLI    & 69.10  & 69.10  & \textbf{69.32} & 59.94 & 59.89          & \textbf{60.47} \\
    SST-2   & 68.99 & 68.86 & \textbf{69.12} & 60.12 & \textbf{60.14} & 59.98          \\
    CoNLL05 & 68.77 & 69.12 & \textbf{69.63} & 59.92 & 59.86          & \textbf{60.22} \\
    \bottomrule
    \end{tabular}
    }
    \caption{Average performance (\%) of all target tasks based on different knowledge distillation losses. Note that ``KL'' denotes the KL-divergence, ``CE'' means the cross-entropy.}
    \label{tab:appendix_loss}
\end{table}

\begin{table*}[t]
    \centering
    \setlength{\tabcolsep}{14pt}
    \scalebox{1}{
    \begin{tabular}{cccccccccc}
    \toprule
    \textbf{Number} & \textbf{CB} & \textbf{COPA} & \textbf{WSC} & \textbf{RTE} & \textbf{WIC} & \textbf{CoLA} & \textbf{MRPC} & \textbf{STSB} & \textbf{CoNLL04} \\ \midrule
    50              & 0.819       & 0.479         & 0.76         & 0.897        & 0.323        & 0.302         & 0.911         & 0.912         & 0.008            \\
    100             & 0.757       & 0.446         & 0.733        & 0.843        & 0.365        & 0.349         & 0.847         & 0.857         & 0.005            \\
    200             & 0.771       & 0.461         & 0.737        & 0.850         & 0.352        & 0.320          & 0.859         & 0.857         & 0.004            \\
    300             & 0.778       & 0.466         & 0.742        & 0.864        & 0.332        & 0.306         & 0.866         & 0.852         & 0.006   \\
    \bottomrule
    \end{tabular}
    }
    \caption{Similarities between the source and target tasks calculated by our metric based on different sampling numbers. Note that we use the QNLI as the source task in this experiment.}
    \label{tab:appendix_sample1}
\end{table*}

\paragraph{\textbf{Sensitivity analysis of our proposed metric on the sampling numbers}}
\label{ablation_sample}
As stated in Section~\ref{sec_panda}, we sample parts of instances from source and target datasets to calculate the similarity. In practice, we sample the data in the dev set for each task, \textit{i.e.}, the sampling number is determined by the minimum number of data in the dev set among all tasks. Specifically, since the dev set of COPA (one lower-resource task) only contains 100 samples, we set 100 as the sampling number. To examine whether our metric is sensitive to the sampling numbers, we calculate the similarity with different sampling numbers. In practice, for the larger sampling numbers, \textit{e.g.}, 200, we sample the data from the dev set for the higher-resource tasks (whose dev sets contain more than 200 instances), and sample the data from the training set or total dataset for the lower-resource tasks. 

\begin{table}[h]
        \centering
        \setlength{\tabcolsep}{16pt}
        \scalebox{1}{
        \begin{tabular}{lcc}
        \toprule
        \zqh \bf Source tasks & \zqh \bf Standard deviation     &\zqh \bf | ours-avg | \\ \midrule
        \zqh MNLI         & \zqh 0.00770 & \zqh 0.00761   \\
        \zqh QNLI         & \zqh 0.01680 & \zqh 0.01783   \\
        \zqh SST-2        & \zqh 0.00455 & \zqh 0.00344   \\
        \zqh CoNLL05      & \zqh 0.00148 & \zqh 0.00158  \\
        \bottomrule
        \end{tabular}
        }
        \caption{{\zqh Standard deviation of predicted similarities across different sampling numbers (\textit{i.e.}, 50, 100, 200, 300 in Table~\ref{tab:appendix_sample1}). ``| ours-avg |'' denotes the absolute difference (average score) between the predicted similarities used in this paper and the average similarities of multiple sampling numbers. BERT-base is used in this analysis.}}
        \label{tab:sample2}
\end{table}

Table~\ref{tab:appendix_sample1} shows the similarities between the source task (QNLI) and target tasks across different sampling numbers. Note that BERT-base is used in this study. It can be seen that the similarities across different sampling numbers are slightly changed. There are relatively larger changes when we only sample 50 instances to calculate the similarity. One of the possible reasons is that too few samples are difficult to truly reflect the characteristics of the task. This can be further confirmed by the observation that as the sampling number increases (higher than 50), the similarities are almost unchanged.

{\zqh Additionally, we report the statistical analysis results of our predicted similarities across different sampling numbers in Table~\ref{tab:sample2}. Note that ``std'' means the standard deviation (average score of all target tasks), and ``| ours-avg |'' denotes the absolute difference (average score) between the predicted similarities in our paper and the average similarities of multiple sampling numbers. It can be seen that the standard deviation of similarities among different numbers is minimal, confirming that our metric is insensitive to the number of samples.}

\subsection{Exploring the Metrics}
\label{sec:discussion}
In addition to the above analysis on \our approach, we also explore our proposed metric and other prior metrics in detail. Specifically, to the best of our knowledge, there are two existing prior metrics about measuring prompt transferability: 1) the metric (namely E$_{avg}$) in SPoT~\cite{vu2021spot} that computes the cosine similarity of average soft prompt embeddings; 2) the other metric~\cite{su2021transferability} (namely ON) that uses the overlapping rate of activated neurons in the feed-forward network as a similarity of prompts.

\begin{table}[t]
    \centering
    \setlength{\tabcolsep}{14pt}
    \scalebox{1}{
    \begin{tabular}{lcccc}
    \toprule
    \multirow{2}{*}{\textbf{Metric}} &
     \multicolumn{2}{c}{\textbf{BERT-large}}              &
     \multicolumn{2}{c}{\textbf{BERT-base}}               \\ 
    \cmidrule(lr){2-3} \cmidrule(lr){4-5}
                            &
     Same            &
     Differ.      &
     Same            &
     Differ.      \\ \midrule \midrule
    E$_{avg}$~\cite{vu2021spot}                  &
     19.3                 &
     5.6                 &
     29.4                 &
     7.6                 \\
    ON~\cite{su2021transferability}                      &
     74.9 &
     26.8 &
     64.5 &
     17.3 \\ \midrule
    Ours                    &
     84.0                 &
     17.0                 &
     75.6               &
     13.4                \\ \bottomrule  
    \end{tabular}
    }
    \caption{Average similarity (\%) of same/different types of tasks predicted by our metric and prior metrics. BERT-large and BERT-base are used.}
    \label{tab:metrics}
\end{table}

\begin{table*}[t]
    \centering
    \setlength{\tabcolsep}{8pt}
    \scalebox{1}{
    \begin{tabular}{clcccccccccc}
    \toprule
    \textbf{Model}& \textbf{Method}        & \textbf{CB}            & \textbf{COPA}        & \textbf{WSC}           & \textbf{RTE}           & \textbf{WIC}           & \textbf{CoLA}          & \textbf{MRPC}        & \textbf{STSB}          & \textbf{Avg.}    &\textbf{$\Delta$}      \\  \midrule \midrule
    \multirow{4}{*}{DeBERTa-v2-xlarge~\cite{he2020deberta}} &model-tuning  & 94.2          & 94.0          & \textbf{80.3} & \textbf{93.2} & 73.3          & 71.1          & \textbf{92.0} & \textbf{92.9} & 86.4    &-      \\
    & \zqh{P-Tuning-v2~\cite{liu2021p}} & 94.6          & 95.0          & 76.7          & 88.8          & 73.1          & 71.0          & 91.7        & 91.8          & 85.3     &*     \\
    & vanilla PoT~\cite{vu2021spot}   & \textbf{98.2} & 95.0          & 76.9          & 89.5          & 73.7          & \textbf{71.4} & 90.6        & 91.9          & 85.9    &+0.6      \\ \cmidrule(lr){2-12}
    & \our         & \textbf{98.2} & \textbf{97.0} & 77.9          & 91.7          & \textbf{75.5} & 69.9          & 91.9        & 92.3          & \textbf{86.8} &+\textbf{1.5} \\ \midrule
    \multirow{4}{*}{\zqh OPT-350m~\cite{zhang2022opt}} &\zqh model-tuning &\zqh 91.1 &\zqh 66.0 &\zqh 63.5 &\zqh 68.9 &\zqh 66.4 &\zqh 53.2 &\zqh 82.4 &\zqh 88.7 &\zqh 72.5 &-      \\
    & \zqh{P-Tuning-v2~\cite{liu2021p}} &\zqh 87.5 &\zqh 69.0 &\zqh 64.4 &\zqh 57.8 &\zqh 66.6 &\zqh 54.8 &\zqh 82.6 &\zqh 88.5   &\zqh 71.4     &\zqh *     \\
    & \zqh vanilla PoT~\cite{vu2021spot}   &\zqh 92.9 &\zqh 75.0 &\zqh \bf  66.3 &\zqh 77.3 &\zqh 69.1 &\zqh \bf  55.6 &\zqh 84.6 &\zqh 89.7      &\zqh 76.3    &\zqh+4.9      \\ \cmidrule(lr){2-12}
    & \zqh \our         &\zqh \bf  94.6 &\zqh \bf  77.0 &\zqh \bf   66.3 &\zqh \bf  79.4 &\zqh \bf  69.7 &\zqh  54.3 &\zqh \bf 85.3 &\zqh \bf  89.8   &\zqh \textbf{77.1} &\zqh +\textbf{5.7} \\ 
    \midrule
    \multirow{4}{*}{\minor LLaMA2-7b~\cite{touvron2023llama}} &\minor model-tuning &\minor 94.6 &\minor 86.0 &\minor 65.4 &\minor 83.3 &\minor \bf 73.7 &\minor 61.4 &\minor 86.3 &\minor \bf  90.9 &\minor 80.2 &\minor -      \\
    & \minor{P-Tuning-v2~\cite{liu2021p}} &\minor 87.5 &\minor 75.0 &\minor 65.4 &\minor 81.2 &\minor 67.9 &\minor 62.5 &\minor 84.8 &\minor 86.1   &\minor 76.3     &\minor *     \\
    & \minor vanilla PoT~\cite{vu2021spot}   &\minor \bf 96.4 &\minor \bf 89.0 &\minor   63.5 &\minor 86.3 &\minor 70.2 &\minor   63.0 &\minor 87.8 &\minor 90.3      &\minor 80.8    &\minor+4.5      \\ \cmidrule(lr){2-12}
    & \minor \our         &\minor   94.6 &\minor \bf  89.0 &\minor \bf   68.3 &\minor \bf  87.4 &\minor   69.9 &\minor \bf 63.3 &\minor \bf 88.0 &\minor  90.8   &\minor \textbf{81.4} &\minor +\textbf{5.1} \\ 
    \bottomrule
    \end{tabular}
    }
    \caption{Performance (\%) on DeBERTa-v2-xlarge, OPT-350m and {\minor LLaMA2-7b}. For the vanilla PoT and \our, we use MNLI as the source task. ``$\Delta$'' denotes the performance improvement compared to prompt-tuning. The best results are in \textbf{bold}.}
    \label{tab:appendix_large}
\end{table*}

\paragraph{\textbf{Whether these metrics can distinguish the different task relationships.}} 
Figure~\ref{fig:heatmap} (Left) shows heatmaps of the prompt transferability predicted by different metrics between all 21 tasks. Specifically, these 21 tasks are first clustered into six groups: NLI, Semantic Matching(SM), QA, NER, SRL, and others (details are shown in Table~\ref{tab:dataset_details}).
Then, we sort these tasks by the groups, \textit{i.e.}, several adjacent tasks are in the same group and have larger task similarities. 
We can observe that our metric captures most intuitive task relationships, \textit{i.e.}, similar tasks have larger prompt transferability. 
However, the other metrics fall short in distinguishing the task relationships, except the relationship of the same dataset itself. 
Furthermore, we compare the predicted similarity of these metrics for two trained prompts within the same type of tasks and between different tasks in Table~\ref{tab:metrics}. It can also be found that our metric works better to distinguish the prompts of the same and different tasks. 
\textbf{Takeaway}: \textit{our proposed metric can better distinguish the different task relationships: similar tasks have larger prompt transferability}.

\paragraph{\textbf{Correlation between these metrics and prompt transfer performance}}
\label{discussion_scores}
In Table~\ref{tab:results_large}, we preliminarily prove that there is a better correlation between predicted transferability and PoT performance when our metric is used. To further systematically verify it, we conduct a quantitative analysis in this part.
Following Su~et~al.~\cite{su2021transferability}, we compute the Spearman's rank correlation scores~\cite{spearman1961proof} between ranks of predicted transferability and PoT performance\footnote{To accurately analyze the effect of metrics, we do not use proposed \our approach, but the vanilla PoT here.}. 
In practice, to reduce the negative impact of knowledge forgetting in the vanilla PoT scenario as much as possible, we use the performance at earlier epochs, \textit{i.e.} the first epoch, as the referential PoT performance, instead of using the final transfer performance.
Figure~\ref{fig:heatmap} (Right) shows the results and we can observe that our proposed metric achieves consistent improvements compared with other metrics across all model sizes, indicating that our metric can make a better choice of which source tasks for a given target task. 
\textbf{Takeaway}: \textit{compared to prior metrics, our metric works better to make appropriate choices to which source tasks for a target task}.

\section{Discussion}
\label{sec_discussion}
Here, we explore whether \our can be extended to more complex scenarios
and more baseline methods. Then, we discuss the contributions of our metric, {\zqh and analyze the parameter and training efficiency of \our in detail. Lastly, we conduct a case study for error analysis.}

\begin{figure}[t]
	\centering
	\includegraphics[width=0.45\textwidth]{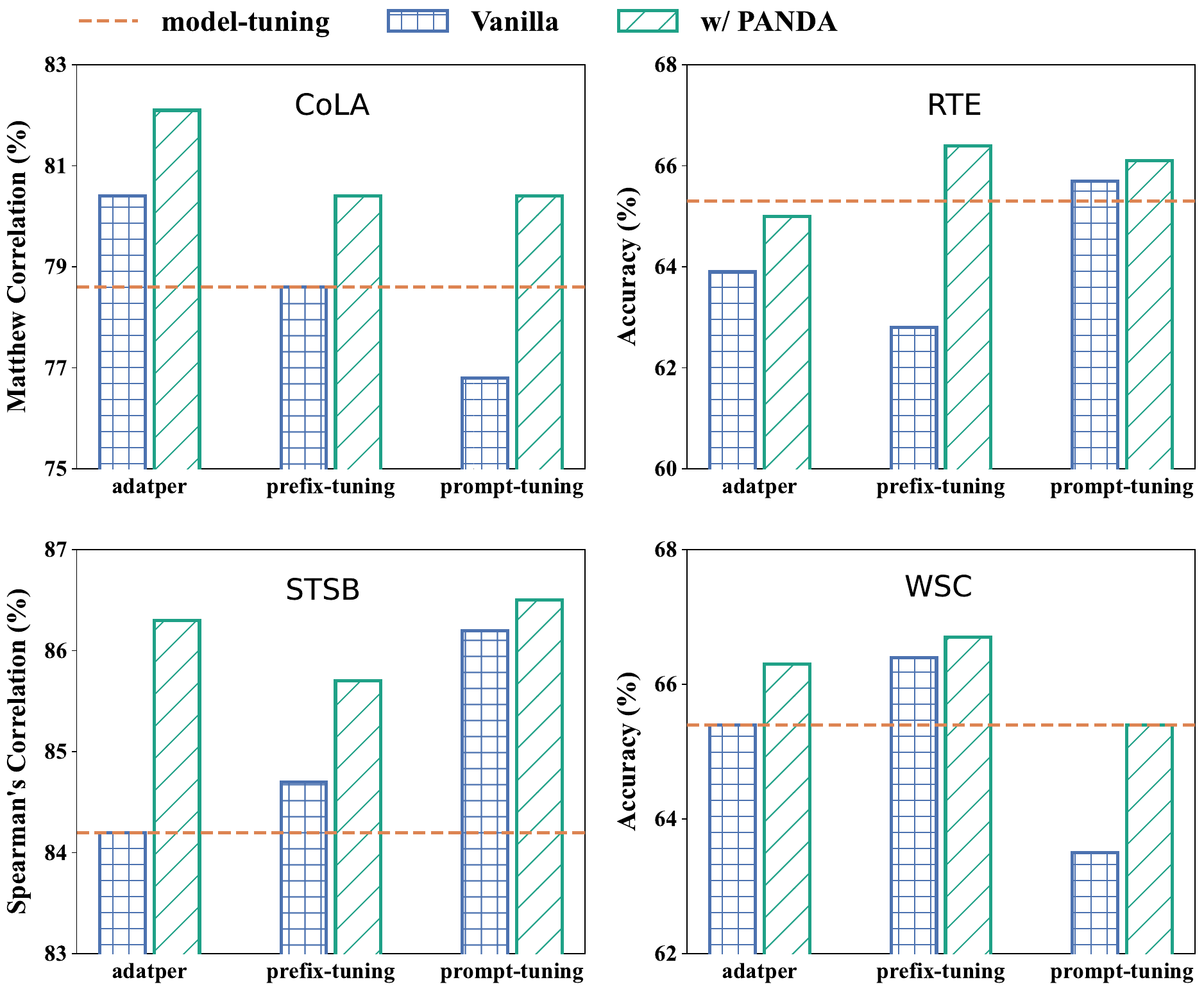} 
	\caption{Complementarity of \our with more baselines. Note that we use MNLI as the source task. BERT-small is used.}
	\label{fig:more_baselines}
\end{figure}

\paragraph{\textbf{Extend \our to more PLM settings}}
\label{appendix_discussion_large_model}
Regarding the question of whether \our can be extended to more PLM settings, 
{\minor \textit{i.e.}, 1) more model architectures and 2) larger model sizes}, we conduct additional experiments as follows. Specifically, for 1), {\minor we additionally evaluate \our on two typical PLMs with different model architectures, \textit{i.e.}, DeBERTa-v2-xlarge~\cite{he2020deberta} (encoder-only PLM) and OPT-350m~\cite{zhang2022opt} (decoder-only PLM).}
Similar to the comparing settings in Section~\ref{main_result}, we compare our \our method to the model-tuning, pure prompt-tuning and vanilla PoT. Note that we use the MNLI as the source task in this experiment and list the detailed results in Table~\ref{tab:appendix_large}. It can be seen that \our outperforms other methods on most tasks, continuing proving the effectiveness of \our.
{\minor On the other hand, since we mainly employ \our on relatively smaller PLMs in above experiments, some readers may concern that whether \our also works well in the currently popular large language models (LLMs). To this end, for 2), we further conduct contrastive experiments on a cutting-edge and powerful LLM, \textit{i.e.}, LLaMA2-7b~\cite{touvron2023llama}, and report the results in Table~\ref{tab:appendix_large} as well.}
{\minor As seen, in the LLM settings, \our can still bring consistent and significant performance gains (up to +5.1\% average score), indicating its universality. These results show that \our has great potential to promote the study of parameter-efficient transfer learning in the currently popular LLM scenarios.}

\begin{figure}[t]
	\centering
	\includegraphics[width=0.45\textwidth]{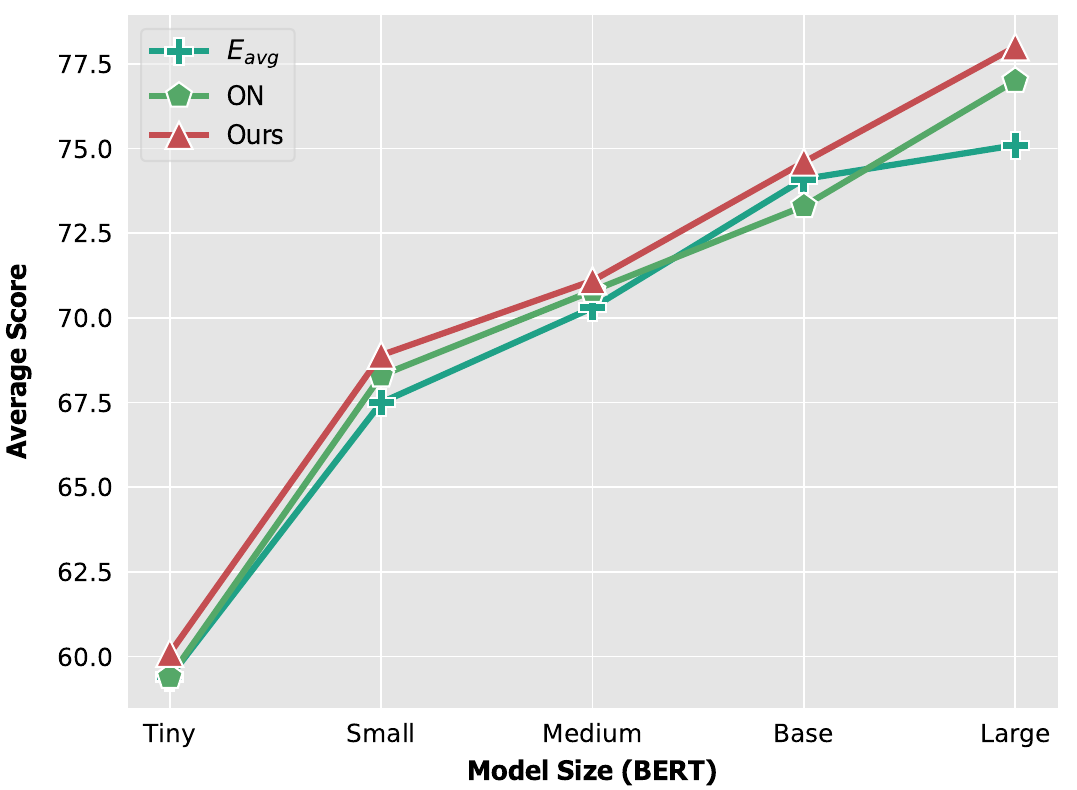} 
	\caption{Average performance (\%) of all target tasks transferred from the most similar source tasks correspondingly, as predicted by different metrics in the vanilla PoT settings.}
	\label{fig:similar_score}
\end{figure}

\paragraph{\textbf{Complementary with other parameter-efficient fine-tuning approaches}}
\label{appendix_discussion_more_baseline}
To further investigate the universality of the framework, we conduct additional experiments on two more parameter-efficient fine-tuning approaches, \textit{i.e.}, adapter~\cite{houlsby2019parameter} and prefix-tuning~\cite{li2021prefix}. Specifically, we show the results of the vanilla approaches and those improved with \our in Figure~\ref{fig:more_baselines}. The results of model-tuning is also shown for reference. Obviously, \our can consistently improve the performance of adapter, prefix-tuning and prompt-tuning by a large margin, \textit{i.e.}, \our is complementary with these approaches as well, which can prove the universality of our \our.  

\begin{table}[]
\centering
\setlength{\tabcolsep}{12pt}
\begin{tabular}{lcc}
\toprule
\zqh \bf Method                  &\zqh  \bf Trainable Parameters &\zqh  \bf Training Time \\ \midrule \midrule
\zqh model-tuning            &\zqh  333.6 M                 &\zqh  566.9 s          \\
\zqh P-Tuning-v2~\cite{liu2021p}           &\zqh  0.98 M                 & \zqh 480.2 s          \\
\zqh vanilla PoT~\cite{vu2021spot} & \zqh 1.9 M                 & \zqh 3790.3 s         \\ \midrule
\zqh \our             & \zqh 1.9 M                 & \zqh 3850.6 s   \\
\bottomrule
\end{tabular}
\caption{{\zqh Comparisons of the number of trainable parameters and training efficiency. Here, we use the STS-B as the target task, and train the BERT-large for 10 epochs. As for PoT methods, we additionally train on the MNLI (source) task for 1 epoch (the main reason for training latency in PoT). All experiments are done with an NVIDIA A100 GPU.}}
\label{tab:latency}
\end{table}

\paragraph{\textbf{Main contributions of our proposed metric}}
\label{appendix_discussion_metric_contribution}
Some readers may point out that the different metrics do not make a big difference in Table~\ref{tab:beta}, and show concerns about the effectiveness of our proposed metric. Here, we discuss the contributions of our metric in detail. Firstly, we state that the main contribution of our metric is to effectively retrieve similar source tasks for a given target task, and the experiment in Table~\ref{tab:beta} is used to prove that the metric can further boost the \our's performance. 
Note that the difference on the BERT-medium is higher than 0.62, which could confirm our statement. 
Secondly, to investigate the effectiveness of our metric intuitively and clearly, we report the performance of the target tasks with the most similar source tasks, as measured by different metrics in the vanilla PoT settings.
Figure~\ref{fig:similar_score} illustrates the results across different PLMs respectively.
It can be seen that our metric consistently achieves the best performance across all model sizes, proving that our metric works better on choosing the useful similar source task for given target task.

{\zqh
\paragraph{\textbf{Parameter quantity and training efficiency of \our}}
To investigate the efficiency of \our, we use BERT-large as the backbone model, setting MNLI and STS-B as the source and target tasks respectively\footnote{{\zqh We set the training epoch as 1 for the source task, and as 10 for the target task. All experiments are done with an NVIDIA A100 GPU in this analysis.}}, and show the comparisons of the number of trainable parameters and training efficiency of different methods in Table~\ref{tab:latency}. As seen, since the training of source prompt introduces some external computations, the training efficiency of \our is relatively lower than the model-tuning. However, \our only has 1.9 M trainable parameters (0.5\% of model-tuning's parameters), indicating its excellent parameter efficiency. Compared to the vanilla PoT, the knowledge distillation process and the training of target-like prompt in \our do not introduce much computations, resulting in small training time increases. In general, \our brings moderate computational budget that is acceptable compared to the performance gain. 
}

\begin{figure}[t]
	\centering
	\includegraphics[width=0.48\textwidth]{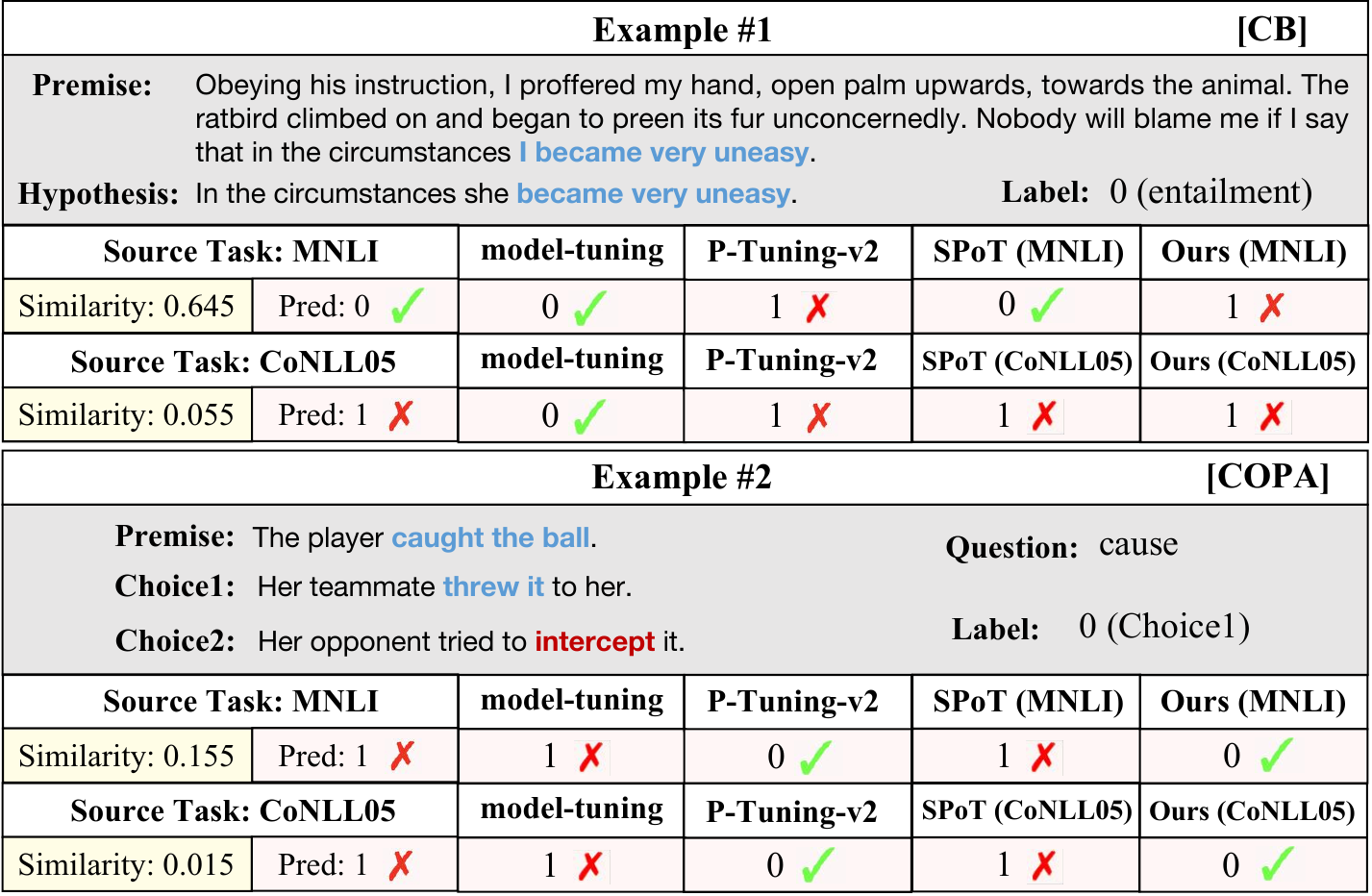} 
	\caption{{\zqh Case study from CB and COPA target tasks. ``Label'' is ground truth label and the numbers in pink cells are the predictions obtained by different models. For the MNLI/CoNLL05 source tasks, we also report the prompt similarities (with the target task) predicted by our metric, and show the individual source model's predictions on these cases.}}
	\label{fig:case_study}
\end{figure}

{\zqh
\paragraph{\textbf{Case Study}}
In this part, we conduct a case study to verify the effectiveness of \our's knowledge transfer ability and reveal its potential limitations in detail. First, we analyze why the vanilla PoT (\textit{i.e.}, SPoT) works well (even outperforms the model-tuning and \our) in some settings, but fails in others. We use the MNLI and CoNLL05 as the source tasks, and use the CB and COPA as the target tasks. The results of BERT-large models are illustrated in Figure~\ref{fig:case_study}. As seen, when the source task (MNLI) is very similar the target task (CB), SPoT performs the excellent performance. However, if the source-target pairs are not similar (\textit{e.g.}, CoNLL05 and COPA), the performance of SPoT would decrease dramatically. These results indicate that SPoT may suffer from the knowledge forgetting problem, confirming our statements.

Second, we would like to see whether \our can effectively transfer the knowledge between the source and target tasks. In the Example \#2 of Figure~\ref{fig:case_study}, it can be seen that, although the individual source models (MNLI and CoNLL05) make the wrong predictions, \our can benefit from their useful general knowledge and help the models predict correctly, showing its effectiveness. However, to be honest, it should be noted that \our could fail in some cases, especially when the source-target pairs are very similar (\textit{e.g.}, MNLI and CB). One of the reasons is that the knowledge distillation process may affect the transfer of useful knowledge. Thus, we believe that exploring a more effective knowledge distillation technique has a great potential to alleviate this limitation and further improve the performance, which is in our future work.
}
\section{Conclusion}
\label{conclusion}
In this paper, we first introduce a new metric to accurately predict the prompt transferability, and then improve PoT with knowledge distillation technique, which uses our metric to adaptively transfer the ``dark knowledge'' from source prompt to the target prompt. Large-scale experiments are conducted to empirically investigate the effectiveness of our methods.
We explore the shortcomings of prior metrics and prove that our metric works better to predict the prompt transferability. Additionally, we show that our \our consistently improves over vanilla PoT by 2.3\% average score across all tasks and models, and makes the prompt-tuning achieve competitive and even better performance than full-parameter model-tuning in various PLM scales scenarios.

\IEEEpeerreviewmaketitle

\ifCLASSOPTIONcaptionsoff
  \newpage
\fi

\bibliographystyle{IEEEtran}
\bibliography{tkde.bib}

\begin{IEEEbiography}[{\includegraphics[width=1in,height=1.25in,clip,keepaspectratio]{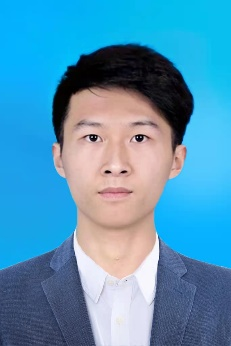}}]{Qihuang Zhong} is currently pursuing the Ph.D. degree in Artificial Intelligence from the School of Computer Science, Wuhan University. His research interests include language model pretraining, natural language understanding and generation. He has authored or co-authored over 10 research papers at top conferences and international journals, including ACL, EMNLP, COLING, IEEE TKDE/TASLP and \textit{etc}. He won the general language understanding (GLUE) and more difficult language understanding (SuperGLUE) challenges.\end{IEEEbiography}

\begin{IEEEbiography}[{\includegraphics[width=1in,height=1.25in,clip,keepaspectratio]{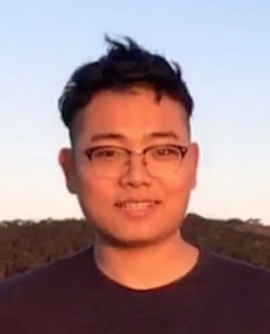}}]{Liang Ding} received Ph.D. from the University of Sydney. He works on deep learning for NLP, including language model pretraining, language understanding, generation, and translation. He published over 40 research papers in NLP/AI, including ACL, EMNLP, ICLR, and ICML. He was the area (session) chair for ACL, AAAI, and SDM.\end{IEEEbiography}

\begin{IEEEbiography}[{\includegraphics[width=1in,height=1.25in,clip,keepaspectratio]{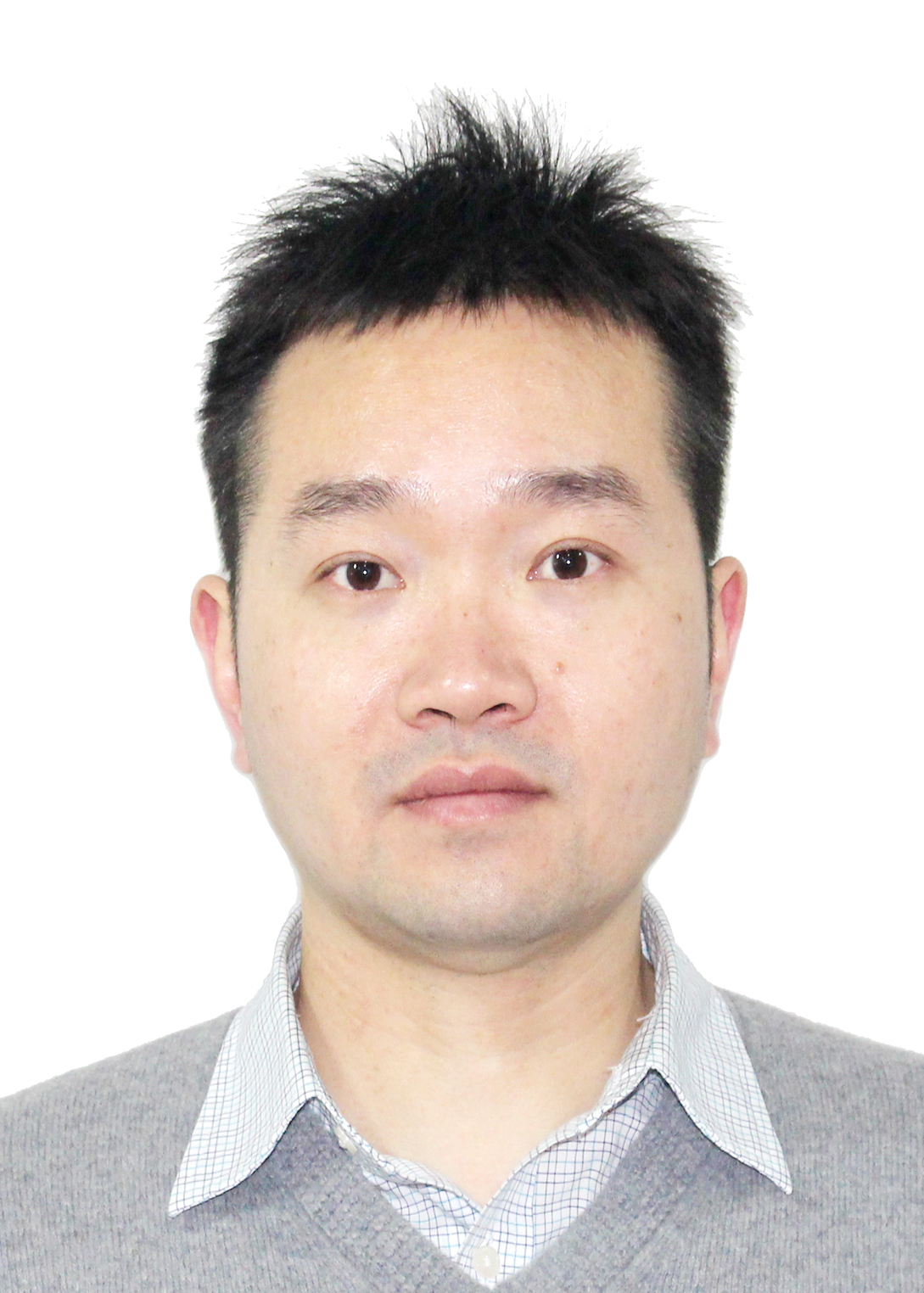}}]{Juhua Liu} is currently a professor with the School of Computer Science and Institute of Artificial Intelligence, Wuhan University. His research interests mainly include image processing, computer vision, natural language processing and machine learning. He has published more than 40 research papers in CV/NLP/AI, including IJCV, IEEE TIP, IEEE TKDE, IEEE/ACM TASLP, CVPR, ACL, AAAI, IJCAI, ACM MM and EMNLP, \textit{etc}. He serves as a reviewer of several top journals, including IEEE TPAMI, IEEE TIP, IEEE TCYB, IEEE TNNLS, IEEE TASLP, IEEE TIM, \textit{etc}, and regularly serves as PC member of CVPR, AAAI, IJCAI, ACM MM, ICASSP and ICME, \textit{etc}. \end{IEEEbiography}

\begin{IEEEbiography}[{\includegraphics[width=1in,height=1.25in,clip,keepaspectratio]{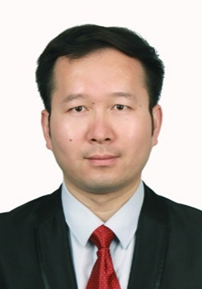}}]{Bo Du} (M'10-SM'15) is currently a professor with the School of Computer Science and Institute of Artificial Intelligence, Wuhan University. He is also the director of National Engineering Research Center for Multimedia Software, Wuhan University, Wuhan, China. He has more than 80 research papers published in the IEEE TPAMI, TIP, IEEE TCYB, IEEE TGRS, \textit{etc}. Fourteen of them are ESI hot papers or highly cited papers. His major research interests include machine learning, computer vision, and image processing. He is currently a senior member of IEEE and serves as associate editor for Neural Networks, Pattern Recognition and Neurocomputing. He won the Highly Cited Researcher (2019\textbackslash2020) by the Web of Science Group. He won IEEE Geoscience and Remote Sensing Society 2020 Transactions Prize Paper Award, the IJCAI (International Joint Conferences on Artificial Intelligence) Distinguished Paper Prize, IEEE Data Fusion Contest Champion, and IEEE Workshop on Hyperspectral Image and Signal Processing Best paper Award.\end{IEEEbiography}

\begin{IEEEbiography}[{\includegraphics[width=1in,height=1.25in,clip,keepaspectratio]{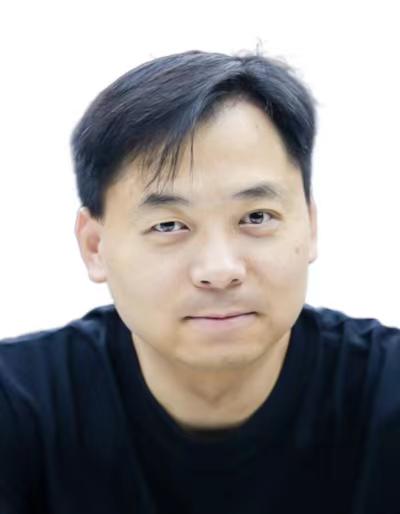}}]{Dacheng Tao} (F'15) is currently a Distinguished University Professor in the College of Computing \& Data Science at Nanyang Technological University. He mainly applies statistics and mathematics to artificial intelligence and data science, and his research is detailed in one monograph and over 200 publications in prestigious journals and proceedings at leading conferences, with best paper awards, best student paper awards, and test-of-time awards. His publications have been cited over 112K times and he has an h-index 160+ in Google Scholar. He received the 2015 and 2020 Australian Eureka Prize, the 2018 IEEE ICDM Research Contributions Award, and the 2021 IEEE Computer Society McCluskey Technical Achievement Award. He is a Fellow of the Australian Academy of Science, AAAS, ACM and IEEE.\end{IEEEbiography}

\appendix
\section{Appendix}

\subsection{Full results of our \our and vanilla PoT approach}
\label{appendix_full_results}
We report all experimental results of our study across model sizes of PLMs. Specifically, in additional the BERT-large's results in Table~\ref{tab:results_large}, we list the full results of BERT-base, BERT-medium, BERT-small and BERT-tiny in Table~\ref{tab:results_base}, Table~\ref{tab:results_medium}, Table~\ref{tab:results_small} and Table~\ref{tab:results_tiny}, respectively. Our PANDA approach achieves consistent and significant performance improvements compared with the vanilla prompt transfer.

\subsection{Prompt transferability predicted by our metric and other metrics}
\label{appendix_prompt_transferability}
\begin{figure*}[ht]
	\centering
	\includegraphics[width=0.95\textwidth]{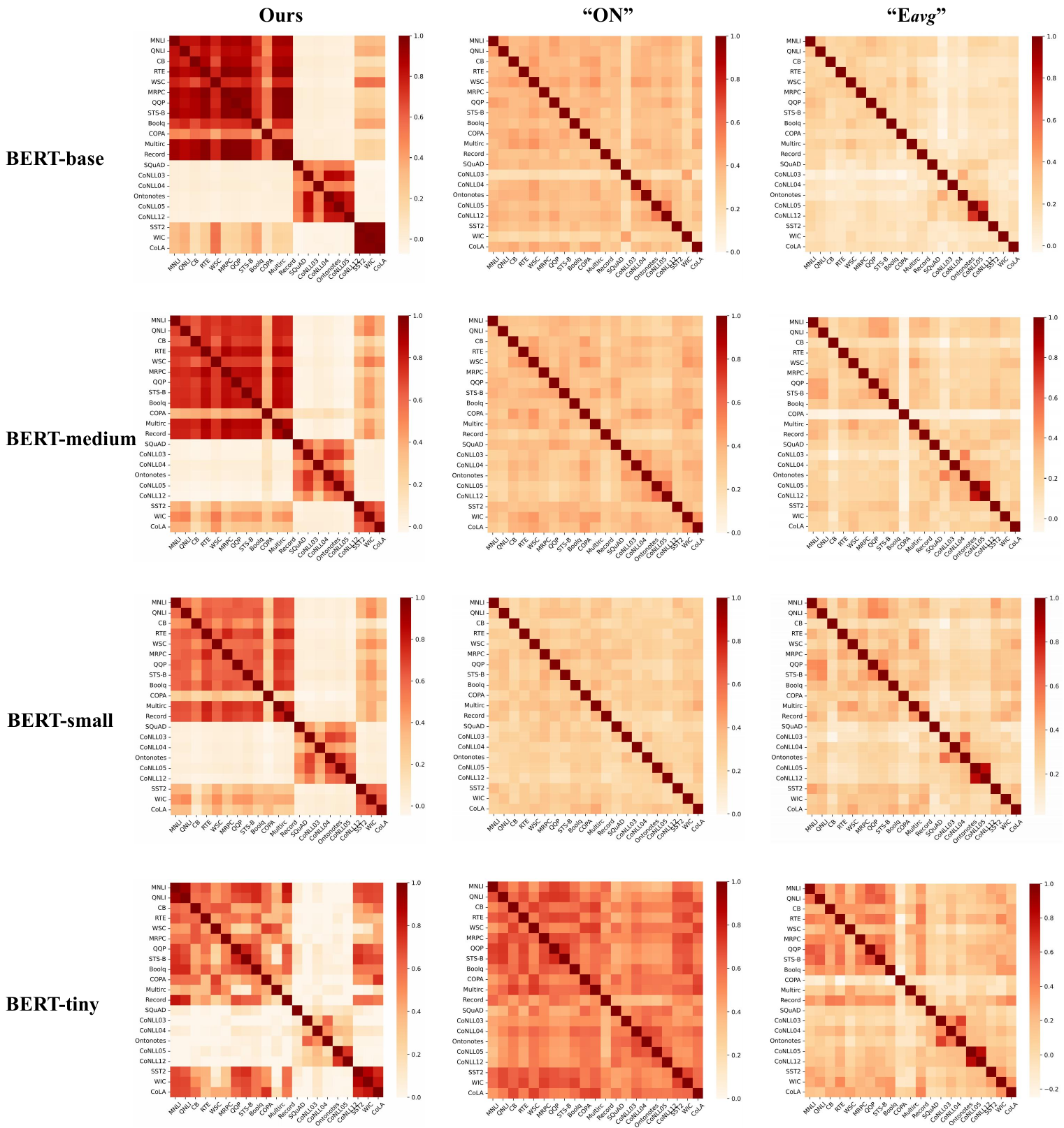} 
	\caption{Comparison of transferability heatmaps predicted by different metrics across all 21 tasks. \textbf{Left}: heatmap of our results; \textbf{Medium}: results predicted by ``ON''; \textbf{Right}: results predicted by ``$E_{avg}$''. We can see that our metric performs better in distinguishing the different task relationships.
	}
	\label{fig:transferability_all}
\end{figure*}

Here, we provide more heatmap results of our predicted prompt transferability across all 21 tasks on different PLMs. Specifically, Figure~\ref{fig:transferability_all} shows the results on BERT-base, BERT-medium, BERT-small and BERT-tiny, respectively. More interestingly, our predicted similarities tend to drop as the scales of PLMs decrease, while the cosine similarities of prompt embeddings show the opposite tendency. The detailed analysis of this phenomenon will be further examined in future work.

\begin{table*}[]
\centering
\setlength{\tabcolsep}{10.8pt}
\scalebox{1}{
\begin{tabular}{lcccccccccl}
\toprule 
\textbf{Tasks} &
 \textbf{CB}   &
 \textbf{COPA} &
 \textbf{WSC}  &
 \textbf{RTE}  &
 \textbf{WIC}  &
 \textbf{CoLA} &
 \textbf{MRPC} &
 \textbf{STSB} &
 \textbf{CoNLL04}   &
 \textbf{Avg.}                         
\\  \midrule \midrule                                            
model-tuning  &
{\color[HTML]{FF0000} \textbf{85.1}}   &
{\color[HTML]{FF0000} \textbf{67.0}}   &
{\color[HTML]{FF0000} \textbf{63.5}}   &
{\color[HTML]{FF0000} \textbf{68.4}}   &
{\color[HTML]{FF0000} \textbf{71.1}}   &
 {\color[HTML]{FF0000} \textbf{54.3}} &
 {\color[HTML]{FF0000} \textbf{89.5}} &
 {\color[HTML]{FF0000} \textbf{88.9}} &
 {\color[HTML]{FF0000} \textbf{85.0}} &
{\color[HTML]{FF0000} \textbf{74.8}}                 

\\				
Lester et al.~\cite{lester2021power}   &
{\color[HTML]{FF9900} \textbf{78.6}}  &
 {\color[HTML]{FF9900} \textbf{67.0}} &
 {\color[HTML]{FF9900} \textbf{63.5}} &
 {\color[HTML]{FF9900} \textbf{61.4}} &
 {\color[HTML]{FF9900} \textbf{61.1}} &
 {\color[HTML]{FF9900} \textbf{35.2}} &
 {\color[HTML]{FF9900} \textbf{75.0}} &
 {\color[HTML]{FF9900} \textbf{86.5}} &
 {\color[HTML]{FF9900} \textbf{79.3}} &
 {\color[HTML]{FF9900} \textbf{67.5}} 

\\

P-Tuning-v2~\cite{liu2021p}   &
{\color[HTML]{FF9900} \textbf{80.4}}  &
 {\color[HTML]{FF9900} \textbf{68.0}} &
 {\color[HTML]{FF9900} \textbf{64.4}} &
 {\color[HTML]{FF9900} \textbf{73.6}} &
 {\color[HTML]{FF9900} \textbf{65.8}} &
 {\color[HTML]{FF9900} \textbf{57.9}} &
 {\color[HTML]{FF9900} \textbf{87.4}} &
 {\color[HTML]{FF9900} \textbf{88.9}} &
 {\color[HTML]{FF9900} \textbf{84.7}} &
 {\color[HTML]{FF9900} \textbf{74.6}}                        
 \\ \midrule
 
\multicolumn{11}{c}{\textbf{\textit{(a) Transfer with Vanilla Prompt Transfer approach~\cite{vu2021spot}. P-Tuning-v2 is used for prompt-tuning in PoT.}}}   \\
\bottomrule 
\end{tabular}
}
\caption{Results (\%) of cross-task prompt transfer on BERT-base. The {\color{red} red-colored} row shows the results of full-tuning BERT-base model, while {\color{orange} orange-colored} ones denote prompt tuning without any prompt transfer. Notably, positive transfers are in {\color[HTML]{32CB00} \bf green} and ``Avg.'' denotes the average performance of all target tasks. Numbers in the subscript indicate relative improvements of \our compared to the vanilla PoT.}
\label{tab:results_base}
\end{table*}
\begin{table*}[]
\centering
\setlength{\tabcolsep}{10.8pt}
\scalebox{1}{
\begin{tabular}{lcccccccccl}
\toprule 
\textbf{Tasks}  &
 \textbf{CB}    &
 \textbf{COPA}  &
 \textbf{WSC}   &
 \textbf{RTE}   &
 \textbf{WIC}   &
 \textbf{CoLA}  &
 \textbf{MRPC}  &
 \textbf{STSB}  &
 \textbf{CoNLL04}   &
 \textbf{Avg.}                         
\\  \midrule \midrule   
model-tuning &
 {\color[HTML]{FF0000} \textbf{81.5}} &
 {\color[HTML]{FF0000} \textbf{66.0}} &
 {\color[HTML]{FF0000} \textbf{65.6}} &
 {\color[HTML]{FF0000} \textbf{63.0}} &
 {\color[HTML]{FF0000} \textbf{66.5}} &
 {\color[HTML]{FF0000} \textbf{42.5}} &
 {\color[HTML]{FF0000} \textbf{86.6}} &
 {\color[HTML]{FF0000} \textbf{86.4}} &
 {\color[HTML]{FF0000} \textbf{81.2}} &
 {\color[HTML]{FF0000} \textbf{71.0}}                         \\
Lester et al.~\cite{lester2021power}   &
{\color[HTML]{FF9900} \textbf{78.6}}  &
 {\color[HTML]{FF9900} \textbf{65.0}} &
 {\color[HTML]{FF9900} \textbf{64.4}} &
 {\color[HTML]{FF9900} \textbf{62.5}} &
 {\color[HTML]{FF9900} \textbf{63.0}} &
 {\color[HTML]{FF9900} \textbf{25.4}} &
 {\color[HTML]{FF9900} \textbf{77.0}} &
 {\color[HTML]{FF9900} \textbf{86.2}} &
 {\color[HTML]{FF9900} \textbf{76.4}} &
 {\color[HTML]{FF9900} \textbf{66.5}} 
 
\\

P-Tuning-v2~\cite{liu2021p}  &
 {\color[HTML]{FF9900} \textbf{80.4}} &
 {\color[HTML]{FF9900} \textbf{65.0}} &
 {\color[HTML]{FF9900} \textbf{63.5}} &
 {\color[HTML]{FF9900} \textbf{67.1}} &
 {\color[HTML]{FF9900} \textbf{64.1}} &
 {\color[HTML]{FF9900} \textbf{43.0}} &
 {\color[HTML]{FF9900} \textbf{84.1}} &
 {\color[HTML]{FF9900} \textbf{87.8}} &
 {\color[HTML]{FF9900} \textbf{79.7}} &
 {\color[HTML]{FF9900} \textbf{70.5}}                         \\
 \midrule
 \multicolumn{11}{c}{\textbf{\textit{(a) Transfer with Vanilla Prompt Transfer approach~\cite{vu2021spot}. P-Tuning-v2 is used for prompt-tuning in PoT.}}}   \\
\bottomrule  
\end{tabular}
}
\caption{Results (\%) of cross-task prompt transfer on BERT-medium. The {\color{red} red-colored} row shows the results of full-tuning BERT-medium model, while {\color{orange} orange-colored} ones denote prompt tuning without any prompt transfer. Notably, positive transfers are in {\color[HTML]{32CB00} \bf green} and ``Avg.'' denotes the average performance of all target tasks. Numbers in the subscript indicate relative improvements of \our compared to the vanilla PoT.}
\label{tab:results_medium}
\end{table*}
\begin{table*}[]
\centering
\setlength{\tabcolsep}{10.8pt}
\scalebox{1}{
\begin{tabular}{lcccccccccl}
\toprule 
\textbf{Tasks}  &
 \textbf{CB}    &
 \textbf{COPA}  &
 \textbf{WSC}   &
 \textbf{RTE}   &
 \textbf{WIC}   &
 \textbf{CoLA}  &
 \textbf{MRPC}  &
 \textbf{STSB}  &
 \textbf{CoNLL04}   &
 \textbf{Avg.}                         
\\  \midrule \midrule   
model-tuning  &
 {\color[HTML]{FF0000} \textbf{78.6}} &
 {\color[HTML]{FF0000} \textbf{66.0}} &
 {\color[HTML]{FF0000} \textbf{65.4}} &
 {\color[HTML]{FF0000} \textbf{65.3}} &
 {\color[HTML]{FF0000} \textbf{62.9}} &
 {\color[HTML]{FF0000} \textbf{38.4}}       &
 {\color[HTML]{FF0000} \textbf{84.5}}       &
 {\color[HTML]{FF0000} \textbf{84.2}}       &
 {\color[HTML]{FF0000} \textbf{72.8}} &
 {\color[HTML]{FF0000} \textbf{68.7}}
\\

Lester et al.~\cite{lester2021power}   &
{\color[HTML]{FF9900} \textbf{76.8}}  &
 {\color[HTML]{FF9900} \textbf{60.0}} &
 {\color[HTML]{FF9900} \textbf{63.5}} &
 {\color[HTML]{FF9900} \textbf{63.2}} &
 {\color[HTML]{FF9900} \textbf{61.3}} &
 {\color[HTML]{FF9900} \textbf{4.6}} &
 {\color[HTML]{FF9900} \textbf{75.0}} &
 {\color[HTML]{FF9900} \textbf{84.3}} &
 {\color[HTML]{FF9900} \textbf{74.2}} &
 {\color[HTML]{FF9900} \textbf{62.5}} 
 
\\

P-Tuning-v2~\cite{liu2021p}   &
 {\color[HTML]{FF9900} \textbf{76.8}} &
 {\color[HTML]{FF9900} \textbf{64.0}} &
 {\color[HTML]{FF9900} \textbf{63.5}} &
 {\color[HTML]{FF9900} \textbf{65.7}} &
 {\color[HTML]{FF9900} \textbf{63.3}} &
 {\color[HTML]{FF9900} \textbf{30.7}} &
 {\color[HTML]{FF9900} \textbf{81.6}} &
 {\color[HTML]{FF9900} \textbf{86.5}} &
 {\color[HTML]{FF9900} \textbf{75.6}} &
 {\color[HTML]{FF9900} \textbf{67.5}}
 \\ \midrule

\multicolumn{11}{c}{\textbf{\textit{(a) Transfer with Vanilla Prompt Transfer approach~\cite{vu2021spot}. P-Tuning-v2 is used for prompt-tuning in PoT.}}}   \\
\bottomrule  
\end{tabular}
}
\caption{Results (\%) of cross-task prompt transfer on BERT-small. The {\color{red} red-colored} row shows the results of full-tuning BERT-small model, while {\color{orange} orange-colored} ones denote prompt tuning without any prompt transfer. Notably, positive transfers are in {\color[HTML]{32CB00} \bf green} and ``Avg.'' denotes the average performance of all target tasks. Numbers in the subscript indicate relative improvements of \our compared to the vanilla PoT.}
\label{tab:results_small}
\end{table*}
\begin{table*}[]
\centering
\setlength{\tabcolsep}{10.8pt}
\scalebox{1}{
\begin{tabular}{lcccccccccl}
\toprule 
\textbf{Tasks}  &
 \textbf{CB}    &
 \textbf{COPA}  &
 \textbf{WSC}   &
 \textbf{RTE}   &
 \textbf{WIC}   &
 \textbf{CoLA}  &
 \textbf{MRPC}  &
 \textbf{STSB}  &
 \textbf{CoNLL04}   &
 \textbf{Avg.}                         
\\  \midrule \midrule   
model-tuning &
 {\color[HTML]{FF0000} \textbf{73.2}} &
 {\color[HTML]{FF0000} \textbf{61.0}} &
 {\color[HTML]{FF0000} \textbf{64.8}} &
 {\color[HTML]{FF0000} \textbf{64.2}} &
 {\color[HTML]{FF0000} \textbf{60.9}} &
 {\color[HTML]{FF0000} \textbf{5.0}} &
 {\color[HTML]{FF0000} \textbf{77.1}} &
 {\color[HTML]{FF0000} \textbf{78.3}} &
 {\color[HTML]{FF0000} \textbf{62.6}} &
 {\color[HTML]{FF0000} \textbf{60.8}} \\
 
 Lester et al.~\cite{lester2021power}   &
{\color[HTML]{FF9900} \textbf{71.4}}  &
 {\color[HTML]{FF9900} \textbf{60.0}} &
 {\color[HTML]{FF9900} \textbf{63.4}} &
 {\color[HTML]{FF9900} \textbf{64.6}} &
 {\color[HTML]{FF9900} \textbf{57.3}} &
 {\color[HTML]{FF9900} \textbf{0.0}} &
 {\color[HTML]{FF9900} \textbf{71.8}} &
 {\color[HTML]{FF9900} \textbf{77.0}} &
 {\color[HTML]{FF9900} \textbf{57.3}} &
 {\color[HTML]{FF9900} \textbf{58.1}} 
 
\\

P-Tuning-v2~\cite{liu2021p}  &
 {\color[HTML]{FF9900} \textbf{71.4}} &
 {\color[HTML]{FF9900} \textbf{62.0}} &
 {\color[HTML]{FF9900} \textbf{63.4}} &
 {\color[HTML]{FF9900} \textbf{65.0}} &
 {\color[HTML]{FF9900} \textbf{57.7}} &
 {\color[HTML]{FF9900} \textbf{0.0}} &
 {\color[HTML]{FF9900} \textbf{72.8}} &
 {\color[HTML]{FF9900} \textbf{81.8}} &
 {\color[HTML]{FF9900} \textbf{57.6}} &
 {\color[HTML]{FF9900} \textbf{59.1}} \\
 \midrule
 \multicolumn{11}{c}{\textbf{\textit{(a) Transfer with Vanilla Prompt Transfer approach~\cite{vu2021spot}. P-Tuning-v2 is used for prompt-tuning in PoT.}}}   \\
\bottomrule  
\end{tabular}
}
\caption{Results (\%) of cross-task prompt transfer on BERT-tiny. The {\color{red} red-colored} row shows the results of full-tuning BERT-tiny model, while {\color{orange} orange-colored} ones denote prompt tuning without any prompt transfer. Notably, positive transfers are in {\color[HTML]{32CB00} \bf green} and ``Avg.'' denotes the average performance of all target tasks. Numbers in the subscript indicate relative improvements of \our compared to the vanilla PoT.}
\label{tab:results_tiny}
\end{table*}

\end{document}